\documentclass[10pt,twocolumn,letterpaper]{article}

\usepackage{iccv}
\usepackage{times}
\usepackage{epsfig}
\usepackage{graphicx}
\usepackage{amsmath}
\usepackage{amssymb}
\usepackage{caption}
\usepackage{arydshln}

\usepackage[pagebackref=true,breaklinks=true,letterpaper=true,colorlinks,bookmarks=false]{hyperref}

\usepackage{dsfont}
\usepackage{multirow}
\usepackage{rotating}
\usepackage[inline]{enumitem}

\usepackage[export]{adjustbox}
\usepackage{titling}

\renewcommand{\hat}{\widehat}
\renewcommand{\tilde}{\widetilde}
\newcommand{\depth}{\zeta}

\let\svthefootnote\thefootnote
\newcommand\freefootnote[1]{%
  \let\thefootnote\relax%
  \footnotetext{#1}%
  \let\thefootnote\svthefootnote%
}

\iccvfinalcopy %

\def\iccvPaperID{6418} %

\begin{document}

\title{%
Reference-guided Controllable Inpainting of Neural Radiance Fields 
}

\author{
Ashkan Mirzaei$^\text{1,2*}$~~~~~~~~~~~~Tristan Aumentado-Armstrong$^\text{1,2,4*}$~~~~~~~~~~~~~Marcus A. Brubaker$^\text{1,3,4}$ \\
Jonathan Kelly$^\text{2}$~~~~~~Alex Levinshtein$^\text{1}$~~~~~~Konstantinos G. Derpanis$^\text{1,3,4}$~~~~~~Igor Gilitschenski$^\text{2}$\\
$^\text{1}$Samsung AI Centre Toronto~~$^\text{2}$University of Toronto~~$^\text{3}$York University~~$^\text{4}$Vector Institute for AI\\
{\tt\small \{a.mirzaei,tristan.a\}@partner.samsung.com,~\{jkelly,gilitschenski\}@cs.toronto.edu}\\
{\tt\small \{kosta,mab\}@eecs.yorku.ca,~alex.lev@samsung.com}
}
\date{}

\twocolumn[{%
\renewcommand\twocolumn[1][]{#1}%
\maketitle
}]

\begin{abstract}
   The popularity of Neural Radiance Fields (NeRFs) for view synthesis has led to a desire for NeRF editing tools. Here, we focus on inpainting regions in a view-consistent and controllable manner. In addition to the typical NeRF inputs and masks delineating the unwanted region in each view, we require only a single inpainted view of the scene, i.e., a reference view. We use monocular depth estimators to back-project the inpainted view to the correct 3D positions. Then, via a novel rendering technique, a bilateral solver can construct view-dependent effects in non-reference views, making the inpainted region appear consistent from any view. For non-reference disoccluded regions, which cannot be supervised by the single reference view, we devise a method based on image inpainters to guide both the geometry and appearance. Our approach shows superior performance to NeRF inpainting baselines, with the additional advantage that a user can control the generated scene via a single inpainted image. Please visit our \href{https://ashmrz.github.io/reference-guided-3d}{project page}. \freefootnote{$^*$ Authors contributed equally. }
\end{abstract}

\begin{figure}[t]
  \centering
   \includegraphics[width=1.0\linewidth]{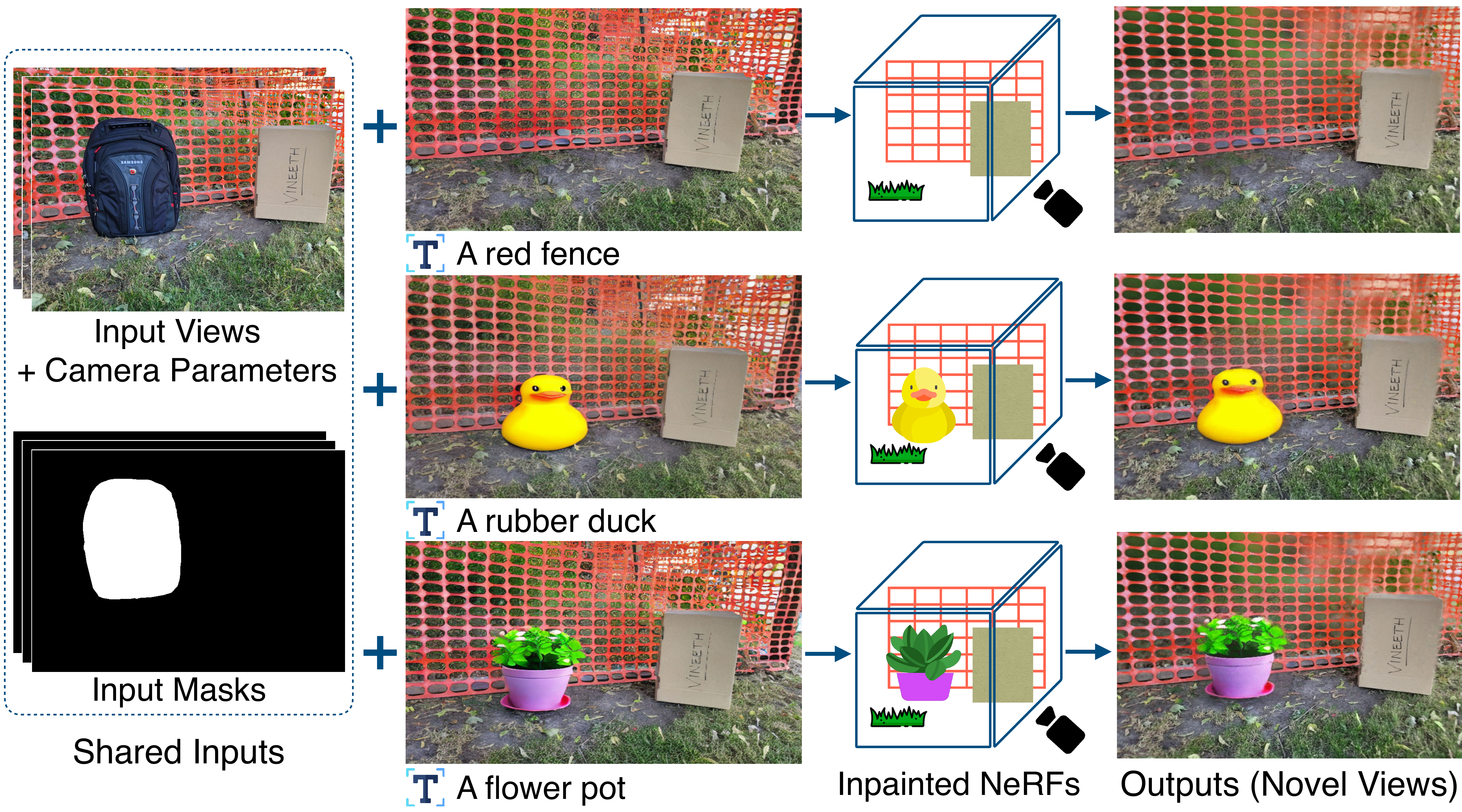}
   \caption{ 
        Visualization of %
        our 3D inpainting approach.
        Starting from (i) a set of posed images (i.e., standard structure-from-motion outputs), 
                      (ii) a multiview mask set associated to (i), and
                      (iii) a \textit{single} inpainted reference image from among (i), 
            we produce a complete inpainted 3D scene, 
            via a novel NeRF fitting algorithm.
        By merely providing a different reference image,
            which can be as simple as changing the text input,
            \textbf{\texttt{T}},
            for a single-image text-conditioned inpainter
            (e.g., \cite{stable.diffusion}), 
            a user can \textit{controllably} generate 3D scenes with the novel desired content.
   }
   \label{fig:overall}
\end{figure}

\section{Introduction}

There has long been intense interest in manipulating  images, due to the broad range of content creation use cases.
Object removal and insertion, corresponding to the image inpainting task, is among the most studied manipulations. Current inpainting models are capable of generating perceptually realistic content that 
conforms to the surrounding image.
Yet, these models are limited to single 2D image inputs; 
our goal is to continue progress in applying such models to the manipulation of full \emph{3D scenes}.

The advent of Neural Radiance Fields (NeRFs) has made transforming real 2D photos into realistic 3D representations more accessible.
As algorithmic improvements continue and computational requirements lessen, such 3D representations may become ubiquitous.
We are thus interested in enabling the same manipulations of 3D NeRFs that are available for images, particularly inpainting (see Fig.~\ref{fig:overall}).

\begin{figure}[t]
  \centering
   \includegraphics[width=0.99\linewidth]{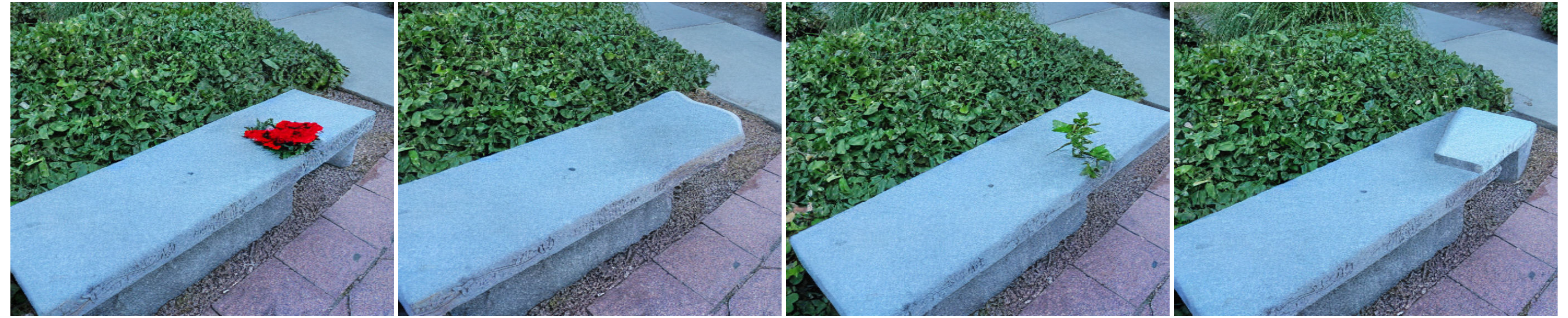}
   \caption{ 
   Sample independent inpaintings \cite{stable.diffusion}
   for four different views of a scene in the SPIn-NeRF dataset~\cite{spinnerf}, using the same prompt. 
   The inpaintings are highly diverse, 
   including some \textit{semantic} differences, not just textural ones.
   }
   \label{fig:different.sd.inpaintings}
\end{figure}

Inpainting in 3D is non-trivial for a number of reasons, such as the paucity of 3D data and the need to account for 3D geometry as well as appearance.
Using NeRFs as a scene representation comes with additional challenges. First, the ``black box'' nature of implicit neural representations makes it infeasible to simply edit the underlying data structure based on geometric understanding. Second, because NeRFs are trained from images, special considerations are required for maintaining multiview consistency. Simply independently inpainting the constituent images using powerful 2D inpainters 
yields viewpoint-inconsistent imagery 
(see Fig.~\ref{fig:different.sd.inpaintings}), 
leading to visually unrealistic outputs.

One approach is to attempt to resolve these inconsistencies \textit{post hoc}. 
For example, NeRF-In~\cite{nerf.in} simply combines views via a pixelwise loss.
More recently, SPIn-NeRF~\cite{spinnerf} improved on this strategy by employing a perceptual loss \cite{perceptual} instead.
Yet, this fails when the inpainted views are perceptually different (i.e., the textures are far apart, even in the perceptual metric space). 
This limits applicability in the case of complex appearances or novel object insertion.
For instance, recent diffusion-based inpainters (e.g., \cite{stable.diffusion,xie2022smartbrush}) can controllably hallucinate novel objects in 2D inpaintings  -- utilizing this capability is currently impossible in the \textit{post hoc} framework. 
In addition, this approach impedes the preservation of specific desired details
(i.e., inter-image conflicts prevent conservation of exact textures). 

In contrast, others have considered \textit{single-reference} inpainting (e.g., \cite{nerf.in}): using only one inpainted view precludes inconsistencies by construction.
However, this lack of 3D information introduces a different set of challenges, including 
(a) poor visual quality in views far from the reference,
in part due to a lack of geometric supervision,
(b) lack of view-dependent effects (VDEs),
and
(c) disocclusions.

In this work, we utilize a single inpainted reference, thus immediately avoiding view inconsistencies, and present a novel algorithm for handling challenges (a-c).  
First, to geometrically supervise the inpainted area, we utilize an optimization-based formulation with monocular depth estimation.
Second, 
    we show how to simulate VDEs
        of non-reference views
        from the reference viewpoint. %
    This enables a guided inpainting approach,
        propagating non-reference colours (with VDEs)
        into the mask. 
Finally, we inpaint disoccluded appearance and geometry 
    in a consistent manner.

We enumerate our contributions as follows:
(i) a single-reference 3D inpainting algorithm
    (depicted in Fig.~\ref{fig:overall}), 
    which avoids visual quality deterioration at views far from the reference;
(ii) a unified method for constructing supervision for masked and disoccluded areas;
(iii) a novel approach to generating VDEs 
    in non-reference views,
    without multiview appearance information;
(iv) significant empirical improvements over prior work, 
    not only in the unprecedented sharpness of novel inpainted views, but also in terms of controllability, enabling users to insert novel objects into 3D scenes by simply providing a single inpainted 2D view.

\section{Related Work}

\textbf{Image Inpainting.} 
Inpainting 2D images has a long %
research history~\cite{jam2021comprehensive,exemplar.based.inpaintint,million.photographs,bertalmio2000image,telea2004image}. 
Neural models represent the state of the art, with advances in
perceptual plausibility~\cite{lama,li2022mat,jain2022keys}, 
multi-scale processing~\cite{glob.local.completion,generative.inpainting,generative.multi.column.conv}, novel architectures~\cite{shiftnet,li2022mat,coherent.sem.attention,lbam}, and generative modelling
(e.g., adversarial~\cite{feature.learning.by.inpainting,comodgan} or
denoising diffusion~\cite{stable.diffusion,lugmayr2022repaint,saharia2022palette,nichol2022glide,luo2023image}). 
To address the ill-posed nature of the inpainting problem, 
pluralistic inpainting methods construct multiple plausible outputs~\cite{comodgan,pluralistic,stable.diffusion,wang2022diverse}. 
Yet, all these methods are 3D unaware. 
In contrast, 3D-aware works are only
partially 3D~\cite{yao20183d}, limited to simple foreground/background scenarios~\cite{3d.photography,jampani2021slide}, 
or cannot synthesize novel views 
of the inpainted result~\cite{zhao2022geofill,transfill}. 
In contrast, we inpaint in an inherently 3D manner via NeRFs, allowing novel-view synthesis of the inpainted scene.

\textbf{NeRF Editing.} 
Neural rendering~\cite{advances.in.neural.rendering} has received significant attention following the success of NeRFs~\cite{original.nerf}, which combines differentiable volumetric rendering~\cite{henzler2019platonicgan,tulsiani2017mvsupervision} and positional encodings~\cite{gehring2017convolutional,vaswani2017attentionisallyouneed,tancik2020fourier}. 
Rapid NeRF developments have improved
visual quality~\cite{ds.nerf,mipnerf,mipnerf.360,bacon,refnerf}, %
fitting or inference speed~\cite{plenoxels,tensorf,instant.ngp,plenoctrees,mobile.nerf,Hedman_2021_ICCV,merf,kurz2022adanerf}, and data requirements~\cite{pixel.nerf,ibrnet,barf,wang2021nerf,poole2022dreamfusion,dreamfields}. 
As NeRFs become more accessible, editing them in 3D has become a topic of interest. 
Recent works provide 3D scene editing capabilities~\cite{paletteNeRF,laterf,clipnerf,yang2021learning,nerf.editing,liu2021editing,conerf,lazova2023control,song2017semantic,dai2021spsg,jheng2022free,li2022compnvs}, 
but either focus on non-inpainting tasks, consider different data availability scenarios, or are limited to simple objects. 
The first NeRF inpainting works are NeRF-In~\cite{nerf.in} and SPIn-NeRF~\cite{spinnerf}. 
Both methods use 2D image inpainters as priors, and fill the unwanted regions of both the training views and the rendered training depths, to guide the generation of the inpainted NeRF. 
While NeRF-In~\cite{nerf.in} 
does not systematically consider the inconsistencies in 
the outputs of 3D-unaware image inpainters (except to reduce the number of reference views),
SPIn-NeRF~\cite{spinnerf} suggests a relaxation based on a perceptual loss~\cite{perceptual} to avoid blur artifacts. 
Although the perceptual loss can handle inconsistencies in the \textit{textures}, it fails if the 2D inpainted views are \textit{semantically} different (e.g., if one inpainted view contains a new object). 
In contrast, our method only relies on a single inpainted view as guidance, while handling VDEs using bilateral solvers~\cite{barron2016fast}. 
This not only enables us to use more powerful image inpainters with greater creative capacity~\cite{stable.diffusion}, but it also allows the user to have more control over the inpainted scene. 
Moreover, we optimize depth and appearance in a unified manner, unlike prior works~\cite{nerf.in,spinnerf}, which treat depth and appearance inpainting separately.

\section{Background: Neural Radiance Fields}

NeRFs~\cite{original.nerf} are an implicit neural field representation (i.e., coordinate mapping) for 3D scenes and objects, generally fit to multiview posed image sets.
The basic constituents are 
        (i) a field, $f_\theta:(x, d) \rightarrow (c, \sigma)$, that maps a 3D coordinate, $x \in \mathbb{R}^3$, and a view direction, $d \in \mathbb{S}^2$, to a colour, $c \in \mathbb{R}^3$, and density, $\sigma \in \mathbb{R}^+$, via learnable parameters $\theta$,
    and (ii) a rendering operator that produces colour and depth for a given view pixel. 
The field, $f_\theta$, can be constructed in a variety of ways (e.g., \cite{original.nerf,plenoxels,bacon,mipnerf.360});
the rendering operator is implemented as the classical volume rendering integral \cite{max2005local}, 
    approximated via quadrature, where
a ray, $r$, is divided into $N$ sections between $t_n$ and $t_f$ (the near and far bounds), with $t_i$ sampled from the $i$-th section. 
The estimated colour is then given by:
\begin{equation}
    \label{eq:volumetric.rendering.discrete}
    \hat{C}(r) = \sum_{i = 1}^{N} T_i(1 - \exp(-\sigma_i \delta_i))c_i,
\end{equation}
where $T_i = \exp(-\sum_{j = 1}^{i - 1} \sigma_j \delta_j)$ is the transmittance, $\delta_i = t_{i + 1} - t_i$, %
and $c_i$ and $\sigma_i$ are the colour and density at $t_i$. %
Replacing $c_i$ with $t_i$ in Eq.~\ref{eq:volumetric.rendering.discrete} 
estimates depth,
$\hat{\depth}(r)$, and disparity (inverse depth), $\hat{D}(r) = \hat{\depth}^{-1}(r)$, instead.

\section{Method}
\label{sec:method}

\begin{figure}[t]
  \centering
   \includegraphics[width=1.0\linewidth]{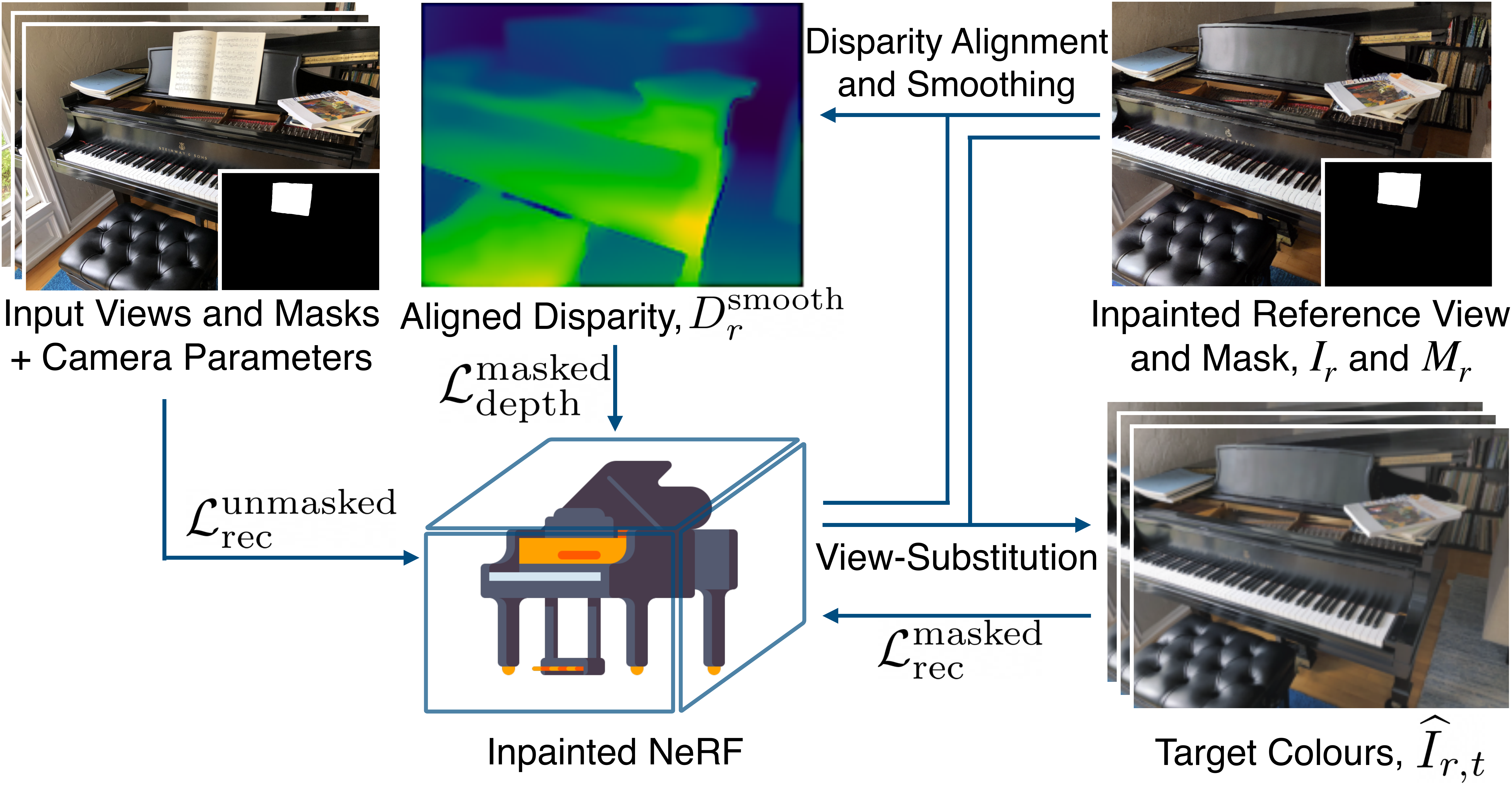}
   \caption{
     Overview of our geometry supervision 
        (\autoref{sec:depth.alignment}) and 
     view-dependent effect (VDE) handling by view-substitution 
        (\autoref{sec:view.sub.full}). 
     Starting from the inpainted reference view, $I_r$, and its mask, $M_r$, (upper-right inset) 
     a disparity map is computed and aligned with the current inpainted NeRF scene (upper-middle inset), 
     then used for masked depth supervision via 
     $\mathcal{L}_\text{depth}^\text{masked}$.
     To supervise view-dependent colours in non-reference views,
        our \textit{view-substitution} technique (\autoref{sec:view.sub}), followed by bilaterally guided inpainting (\autoref{sec:bilateral.inp.residual}),
        obtains target colours ($\widehat{I}_{r,t}$; lower-right inset), used for $\mathcal{L}_\text{rec}^\text{masked}$.
     Finally, \textit{un}masked areas of the NeRF can be supervised with standard losses, $\mathcal{L}_\text{rec}^\text{unmasked}$, from the unmasked inputs
     (leftmost inset). 
     Note that the masked supervisory sources (for $\mathcal{L}_\text{depth}^\text{masked}$ and $\mathcal{L}_\text{rec}^\text{masked}$) 
        are periodically recomputed throughout fitting as the NeRF evolves.\protect\footnotemark
   }
   \label{fig:method}
\end{figure}

\footnotetext{IBRNet images in Fig. 3,5,9 by Wang et al. available in \href{https://drive.google.com/drive/folders/1qfcPffMy8-rmZjbapLAtdrKwg3AV-NJe}{IBRNet}~\cite{ibrnet} 
under a \href{https://creativecommons.org/licenses/by/3.0}{CC BY 3.0 License}.}

The inputs in our setup are $n$ input images, $\{ I_i \}_{i=1}^{n}$, their camera transform matrices, $\{ \Pi_i \}_{i=1}^{n}$, and their corresponding masks, $\{ M_i \}_{i=1}^{n}$\footnote{We assume that masks are given, but they can be obtained automatically with interactive 3D segmentation methods~\cite{spinnerf,neural.object.selection}.}, delineating the unwanted region.
We assume a \textit{single} inpainted reference view, $I_r$, where $r \in \{1, 2, \ldots, N \}$, 
which provides the information that a user expects to be extrapolated into a 3D inpainting of the scene.
We propose an approach to use $I_r$, 
    not only to inpaint the NeRF, 
    but also to generate 3D details and VDEs from other viewpoints as well.
In~\autoref{sec:depth.alignment}, we introduce the use of monocular depth estimators to guide the geometry of the inpainted region, according to the depth of the reference image, $I_r$. 
In~\autoref{sec:view.sub.full}, we propose the use of bilateral solvers~\cite{barron2016fast}, in conjunction with our view-substitution technique, to add VDEs to views other than the reference view. 
See Fig.~\ref{fig:method} for a depiction of our 
geometry supervision and VDE handling.
Since not all the masked target pixels are visible in the reference, in~\autoref{sec:disoccluded.regions},
    we devise an approach to provide supervision for such disoccluded pixels, via additional inpaintings.

\textbf{Complete Loss Function.}
Our overall objective for inpainted NeRF fitting is given by: 
\begin{align}
    \label{eq:full.loss}
    \mathcal{L} =&
\mathcal{L}_\text{rec}^\text{unmasked} + 
\gamma_\text{depth}^\text{masked} \mathcal{L}_\text{depth}^\text{masked} + 
\gamma_\text{rec}^\text{masked} \mathcal{L}_\text{rec}^\text{masked} + 
\gamma_\text{do}\mathcal{L}_\text{do},
\end{align} 
where $\mathcal{L}_\text{rec}^\text{unmasked}$, 
$\mathcal{L}_\text{depth}^\text{masked}$, 
$\mathcal{L}_\text{rec}^\text{masked}$, and 
$\mathcal{L}_\text{do}$
represent the unmasked appearance loss,
masked geometry loss,
view-dependent masked colour loss,
and disocclusion loss, respectively
(detailed below).
The latter three loss terms have weights $\gamma_\text{depth}^\text{masked}$, $\gamma_\text{rec}^\text{masked}$, and $\gamma_\mathrm{do}$. 
Note that 
the supervision for 
$\mathcal{L}_\text{depth}^\text{masked}$, $\mathcal{L}_\text{rec}^\text{masked}$, and $\mathcal{L}_\text{do}$ 
are computed every $N_\text{depth}$, $N_\text{bilateral}$, and $N_\text{do}$ iterations (and hence those losses are not utilized until that many iterations have passed).

\subsection{Supervising Reference View Geometry} 
\label{sec:depth.alignment}

In the first stage of our algorithm, $f_\theta$ is supervised on the \textit{un}masked pixels for $N_\text{depth}$ iterations, via the standard NeRF reconstruction loss:
\begin{equation}
\label{eq:reconstruction.loss.unmasked}
\mathcal{L}_\text{rec}^\text{unmasked} = \mathbb{E}_{r \sim \mathcal{R}_\text{unmasked}} \big\Vert \hat{C}(r) - C_{\text{GT}}(r) \big\Vert ^2,
\end{equation}
where $\mathcal{R}_\text{unmasked}$ is the set of rays corresponding to the pixels in $\{ I_i \odot (1 - M_i)\}_{i=1}^{n}$, and $C_{\text{GT}}(r)$ is the GT colour for the ray, $r$. 
As a result, while the geometry and appearance of the unmasked parts of the scene begin to converge, the masked region remains under-fit (the masked area is not fit via $I_r$ at this point, as it makes altering masked values in later stages more difficult). 
The only available guidance for such masked pixels resides in $I_r$; however, this only provides single-view appearance information, which cannot directly contribute geometric supervision. 

\textbf{Masked Reference Disparity}.\ 
To address this challenge, we propose the use of a monocular depth estimator~\cite{midas1, midas2}, $\tilde{D}(\cdot)$, to predict the uncalibrated disparity of the source view, $\tilde{D}_r = \tilde{D}(I_r)$, and guide the geometry. 
However, the predicted reference depth, $\tilde{D}_r^{-1}$, 
is non-metric, resides in a different coordinate system, and may be inaccurate, as it was predicted from a single frame.
As a result, before supervising the disparity of the NeRF using $\tilde{D}_r \odot M_r$, we need to align $\tilde{D}_r$ to our rendered NeRF reference disparity, $\hat{D}_r$. %
Although under-fit on the masked pixels, $\hat{D}_r$ has reliable values for the \textit{un}masked pixels. 

However, not all of the masked pixels are equally important: areas close to the mask boundary need to be tightly aligned, to ensure the mask edge is minimally visible in the final results, whereas it is not critical to completely align the depths far from the mask, since only the masked pixels will receive supervision with the aligned reference disparity.
Thus, we propose to align $\tilde{D}_r$ and $\hat{D}_r$ for the reference view on the unmasked pixels in a weighted manner, giving higher weight to the points closer to the mask. 

\textbf{Weighted Disparity Alignment}.\ 
Traditionally, a scale $a_0$ and an offset $a_1$ are used to affinely transform 2.5D disparity maps, $\tilde D_r$, to $a_0 \tilde D_r + a_1 \mathds{1}_{HW}$~\cite{midas1}, where $H$ and $W$ are the height and width of the input images, and $\mathds{1}_{HW}$ is an $H\times W$ all-one matrix. We further increase the degrees of freedom of the alignment to have a tighter alignment around the mask edges. 
We use two additional $H\times W$ matrices, $\mathcal{H}$ and $\mathcal{V}$, 
where for a pixel $p = (p_x, p_y)$, $\mathcal{H}(p) = p_x$ and $\mathcal{V}(p) = p_y$. 
Intuitively, 
these enable additional axis-aligned ``tilts'' 
that improve fitting quality.
Please see our supplementary material for an illustration of $\mathcal{H}$ and $\mathcal{V}$. 
Then, the aligned predicted inverse depth is:
\begin{equation}
    \label{eq:aligned.depth}
    D_r = a_0 \tilde D_r + a_1 \mathds{1}_{HW} + a_2 \mathcal{H} + a_3 \mathcal{V}.
\end{equation}
Since the pixels closer to the mask are more important for our inpainting application, we use the following weighted objective function to solve for the scalars, $a_i$: %
\begin{equation}
    \label{eq:weighted.objective}
    F_\mathrm{wf}(\{ a_i\}_i) = 
    \sum_{p \in I_r \odot (1 - M_r)} w(p)
    \left[ %
    D_r(p) - \hat D_r(p) 
    \right]^2 %
\end{equation}
where $p$ is an unmasked pixel from the source view, and $w(p)$ is the weight of $p$, which is the inverse of the distance between $p$ and the 
mask centre-of-mass. 

While $D_r$ has significantly improved alignment, misalignments still tend to persist near the edges of $M_r$. 
We thus conduct an additional optimization step, where we correct $D_r$ to encourage greater smoothness around the mask, yielding $D_r^\text{smooth}$ (details in the supplementary material).

\textbf{Loss}. After alignment and smoothing, $D_r^\text{smooth}\odot M_r$ supervises the masked region of the reference view, $I_r$,
via:
\begin{equation}
    \label{eq:depth.loss}
    \mathcal{L}_\text{depth}^\text{masked} = \mathbb{E}_{r' \in \mathcal{R}_\text{masked}} 
    \left[
    \hat D_r(r') - D_r^\text{smooth}(r') 
    \right]^2 %
\end{equation}
Note that
$D_r^\text{smooth}$, is recalculated every $N_\text{depth}$ iterations to utilize the latest fitted geometry, $\hat D_r$.

\subsection{View-dependent Effects by View-substitution} \label{sec:view.sub.full}

Now that the inpainted region is being geometrically supervised by the depth loss, $\mathcal{L}_\text{depth}^\text{masked}$, 
    we can also supervise the NeRF appearance in the masked region with $I_r$ (see \autoref{sec:supervision.from.ref}).
    Here, we detach the gradients of the densities to prevent the colour loss from affecting the geometry.
However, supervision %
within the masked region from $I_r$ alone does not account for view-dependent changes
(e.g., specularities and non-Lambertian effects).
To correct this, we propose an approach 
    that enables adding view-dependent effects (VDEs) to the masked area from non-reference viewpoints,
    by correcting reference colours to match the surrounding context of the other views.

In this section, we consider a target view, $I_t\in \{ I_i\}_{i=1}^n$. 
First, in~\autoref{sec:view.sub}, we propose our \textit{view-substitution} method, to enable rendering the scene from the reference camera, but with the colours of $I_t$. 
Then, we inpaint the residual between this target-colour render and $I_r$,
    propagating the image context of $I_t$ into the masked area, including VDEs (\autoref{sec:bilateral.inp.residual}).
Finally, this residual is applied to obtain corrected reference colours, 
    which are used in~\autoref{sec:supervision.from.ref} to supervise the NeRF in masked areas of $I_t$. 

\begin{figure}[t]
  \centering
   \includegraphics[width=1.0\linewidth]{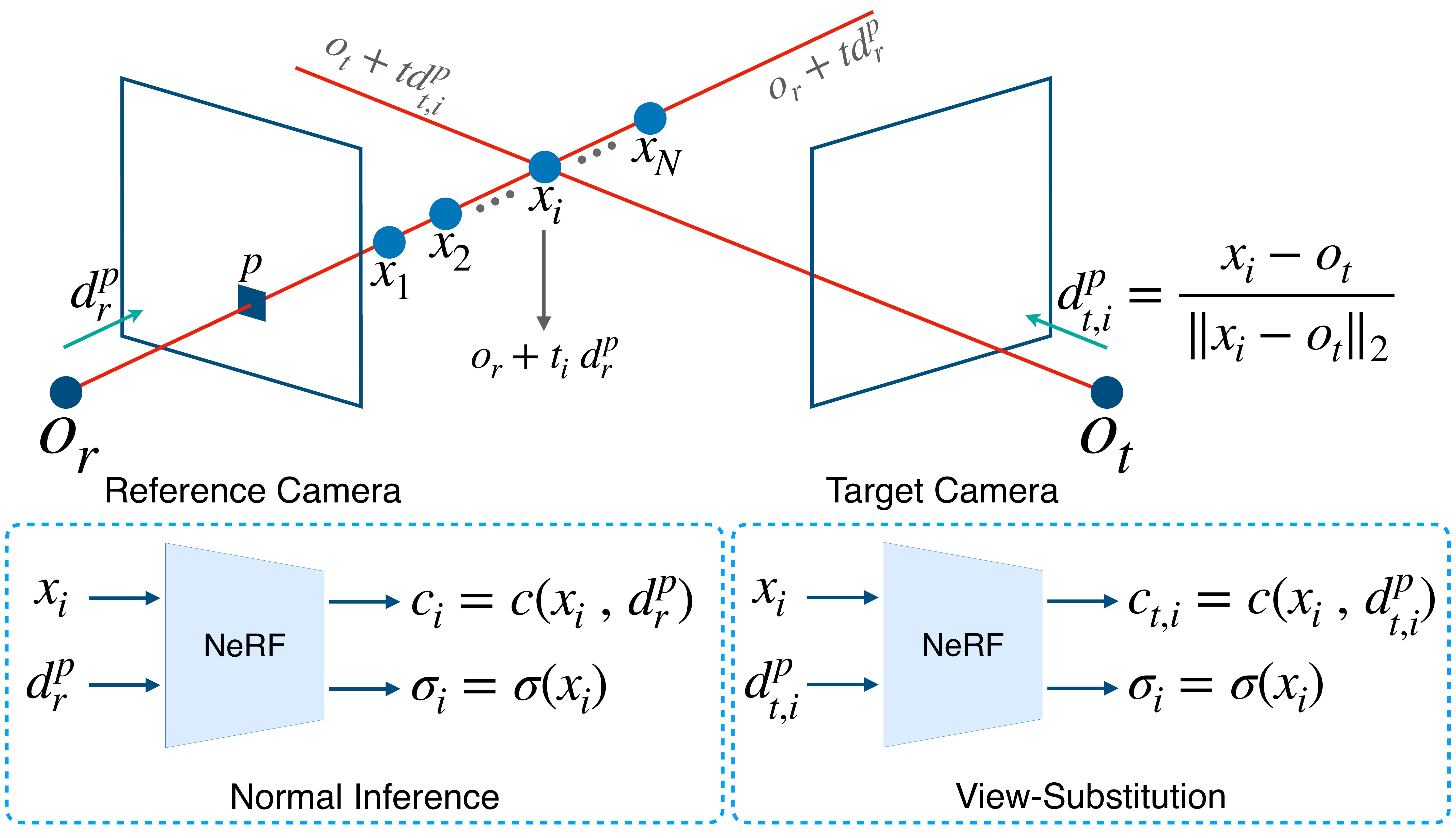}
   \caption{ 
      Depiction of our \textit{view-substitution} technique, which enables rendering from the reference viewpoint, but with the view-dependent effects of a target viewpoint, by simply substituting the directional input to the per-shading-point neural colour field.
      Upper inset: given a shading point position, $x_i$, on a ray emanating from the reference camera (with direction $d_p^r$), 
        we can obtain the corresponding ray direction, $d_{t,i}^p$, that intersects $x_i$ from a target-image camera (at $o_t$).
      Lower inset: (left) standard inputs used to query the NeRF for the colour, $c(x_i, d_r^p)$, at shading point $x_i$;
                   (right) view-substituted inputs used to query the NeRF, obtaining $c(x_i, d_{t,i}^p)$ as the colour instead.
   }
   \label{fig:view.substitution}
\end{figure}

\subsubsection{View-substitution} \label{sec:view.sub}

Our view-substitution technique enables looking at the scene 
    from the reference viewpoint, 
    while changing the shading point colours 
    (i.e., $c_i$ in Eq.~\ref{eq:volumetric.rendering.discrete}) 
    during rendering to have the colours of a target view. 
Intuitively, this allows us to construct multiple ``versions'' of the \textit{reference} view, each with colours corresponding to the VDEs of a \textit{target} view.
    
Fig.~\ref{fig:view.substitution} shows an overview of our view-substitution method.
Consider a pixel, $p$, from the reference view, $I_r$. 
During standard NeRF rendering, a ray is cast through the scene, passing from the camera origin, $o_r$, through the pixel, $p$.  
This ray is parameterized as $x(t) = o_r + td_r^p$, with direction $d_r^p \in \mathbb{S}^2$.
Next, shading points $x_1, x_2, \ldots, x_n \in \mathbb{R}^3$ are sampled on this ray. 
Normally, for the $i$-th sample on the ray, its coordinates, $x_i$, and the view-direction, $d_r^p$, are fed to the NeRF model to obtain the density, $\sigma(x_i)$, and the colour \textit{from the reference viewpoint}, $c(x_i, d_r^p)$. 
However, here, instead of the reference view colours, we are interested in the colours of the points as if they were viewed from the target camera. 
As a result, when computing the shading point colour, we substitute the view direction, $d_r^p$, for the direction acquired by connecting the origin of the target view, $o_t$, and the shading point, $x_i$. This direction is computed via:
\begin{equation}
    \label{eq:view.sub.new.direction}
    d_{t,i}^p = {(x_i - o_t)} / {\Vert x_i - o_t \Vert_2},
\end{equation}
resulting in view-substituted shading point colours $c(x_i, d_{t,i}^p)$ instead.
We can then volume render \textit{from the reference viewpoint} across pixels, but \textit{with the view-substituted target colours}, to obtain rendered images 
$I_{r, t}$. 
Such images have the structure of the reference view (e.g., edges), but the appearance (and thus VDEs) of the target view.
Please see our supplementary material for additional details and visualizations.

\begin{figure}[t]
  \centering
   \includegraphics[width=0.99\linewidth]{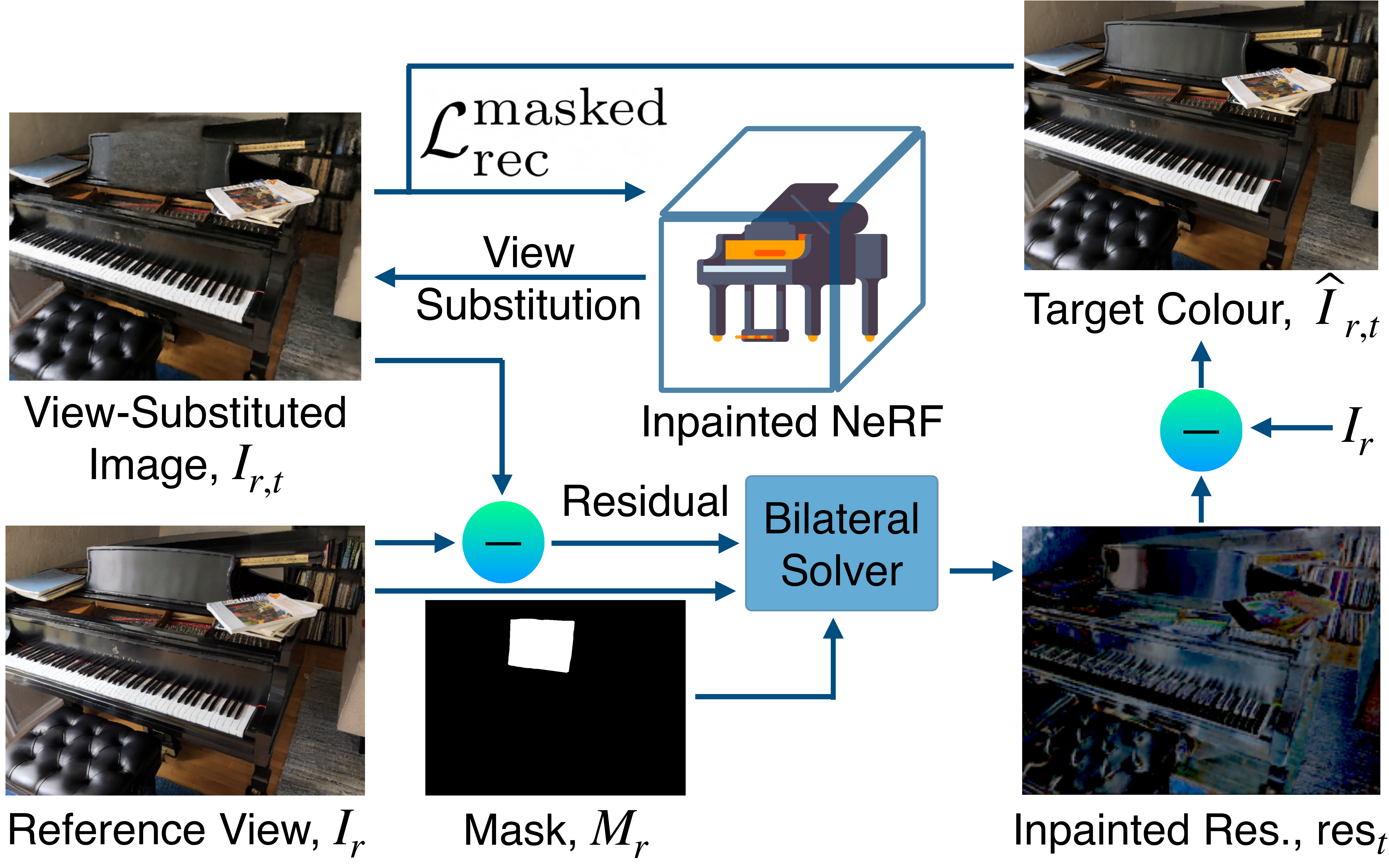}
   \caption{ 
      Overview of our view-dependent effect (VDE) handling approach. 
      For each target, $t$, the scene is rendered from the reference camera with target colours to get the view-substituted image, $I_{r, t}$ (top-left). A bilateral solver inpaints the residual between the reference view and the view-substituted image, resulting in the inpainted residual, $\text{res}_t$ (bottom-right), which is subtracted from the reference view to get the target colour, $\hat I_{r, t}$ (top-right). The discrepancy between the target colours and the view-substituted images provides supervision for the masked region. 
   }
   \label{fig:view.sub.summary}
\end{figure}

\subsubsection{Bilateral Solver for Residuals} \label{sec:bilateral.inp.residual}

At this point, before any supervision on the masked areas of the target images has begun, 
    the view-substituted rendering, $I_{r, t}$, will likely be under-fit inside the mask, $M_r$, 
    but should have meaningful colours outside of the mask (see Fig.~\ref{fig:view.sub.summary}, top-left).
Consider the residual, $\Delta_t = I_r - I_{r, t}$, 
    which measures the difference between the reference and target colours (from the reference viewpoint).
We want to use the values of this residual \textit{out}side the mask to predict plausible values for the residual \textit{in}side of the mask. 
We rely on the assumption that VDEs (encapsulated by the residuals) cannot have high-frequency variations when there is no edge in the reference view, $I_r$. 
In other words, 
    if there is little image contrast in a given region of $I_r$, 
    we only expect smooth changes in the VDEs of $\Delta_t$. 
This is a natural assumption, as changes in materials and reflectance properties are usually accompanied by image edges, demarcating the boundaries between objects or object parts.
The bilateral solver~\cite{barron2016fast}, denoted $\mathcal{B}$, is thus an intuitive approach to inpainting the residual inside the mask, 
    as it enables directly using the edges of $I_r$ for guidance. 
Briefly, $\mathcal{B}$ optimizes an image signal, 
    balancing confidence-weighted reconstruction fidelity 
    and bilateral smoothness, 
    guided by the structure of an additional RGB reference image.
This is analogous to ``diffusing'' in pixel values from outside the mask \cite{barash2002fundamental}, directed by the reference.
In our case, 
    $\mathcal{B}$ thus utilizes $I_r$ as the reference input 
    (from which the edge guidance occurs through the bilateral affinities),
    while using $\Delta_t$ as the target (valid only outside the mask).
We set the confidence to the maximum possible value ($c_\text{max}$) outside of the mask and to zero inside it. 
Then, we run $\mathcal{B}$ to get the inpainted residual:
\begin{equation}
    \label{eq:bilateral.solving.residual}
    \text{res}_t = \mathcal{B}\big(I_r, \;\;I_r - I_{r, t}, \;\;(1 - M_r) \times c_\text{max} \big).
\end{equation}
The target colours are then obtained as $\hat I_{r, t} = I_r - \text{res}_t$. 
Note that $\text{res}_t$ equals $\Delta_t$ \textit{out}side the mask, but we only need its values \textit{in}side the mask for supervision.
To ensure this supervision remains up-to-date with the changing NeRF, every $N_\text{bilateral}$ iterations, we re-render the view-substituted images, run $\mathcal{B}$, and  
compute $\hat I_{r, j}\;\forall\;j\in[1,n]$ (with $\hat I_{r, r} = I_r$).

\subsubsection{Supervision from the Reference View} \label{sec:supervision.from.ref}
Once the view-substituted (\autoref{sec:view.sub}) and bilaterally inpainted (\autoref{sec:bilateral.inp.residual}) target renders, $\{\hat{I}_{r,j}\}_{j=1}^n$, are available (after reaching $N_\text{bilateral}$ at least once),
    we are now able to supervise the masked appearances of the target images.
Note that each such image $\hat{I}_{r,t}$ looks at the scene via the reference source camera (i.e., has the image structure of $I_r$), but has the colours (in particular, VDEs) of $I_t$.
We utilize those colours, obtained by the bilateral solver, to supervise the target view appearance under the mask.
To do so, we render each view-substituted image inside the mask 
(obtaining $I_{r,t}$, as in~\autoref{sec:view.sub}), and compute a reconstruction loss by comparing it to the bilaterally inpainted output, $\hat{I}_{r,t}$:
\begin{equation}
    \label{eq:reconstruction.loss.masked}
    \mathcal{L}_\text{rec}^\text{masked} = \frac{1}{n} \sum_{t=1}^n \mathbb{E}_{r' \sim \mathcal{R}_\text{masked}^r} \big\Vert I_{r,t}(r') - \hat I_{r, t}(r') \big \Vert^2, 
\end{equation}
where $\mathcal{R}_\text{masked}^r$ is the set of rays corresponding to the masked pixels in the reference view ($\mathds{1}_{HW}\odot M_r$). Fig.~\ref{fig:view.sub.summary} provides an overview of our 
VDE-handling component.

\textbf{Filtering Edge Islands.}
Sometimes it is not possible for the bilateral solver, $\mathcal{B}$, to propagate values from outside the mask to certain areas on the inside of it.
This occurs whenever there is an ``edge island'' in the masked region:
i.e., a disconnected area in bilateral space (e.g., see \cite{marki2016bilateral}), 
    such that information from outside the mask will not be transmitted inside.
This typically leads to out-of-distribution values in the output from $\mathcal{B}$. 
Here, our goal is to remove such values from consideration.
Our approach roughly corresponds to imposing a Lambertian prior on object appearance,
to which we default when $\mathcal{B}$ is too uncertain;
in such cases, the target colours will likely end up close to those of the reference view
(though this is not guaranteed, due to the view-dependent MLP). To implement this strategy, we detect and filter out-of-distribution values associated to rays in $\mathcal{R}_\text{masked}^r$, when calculating Eq.~\ref{eq:reconstruction.loss.masked}, from every target view $t \neq r$; see our supplement for details.

\subsection{Disoccluded Regions} \label{sec:disoccluded.regions}

\begin{figure}[t]
  \centering
   \includegraphics[width=0.99\linewidth]{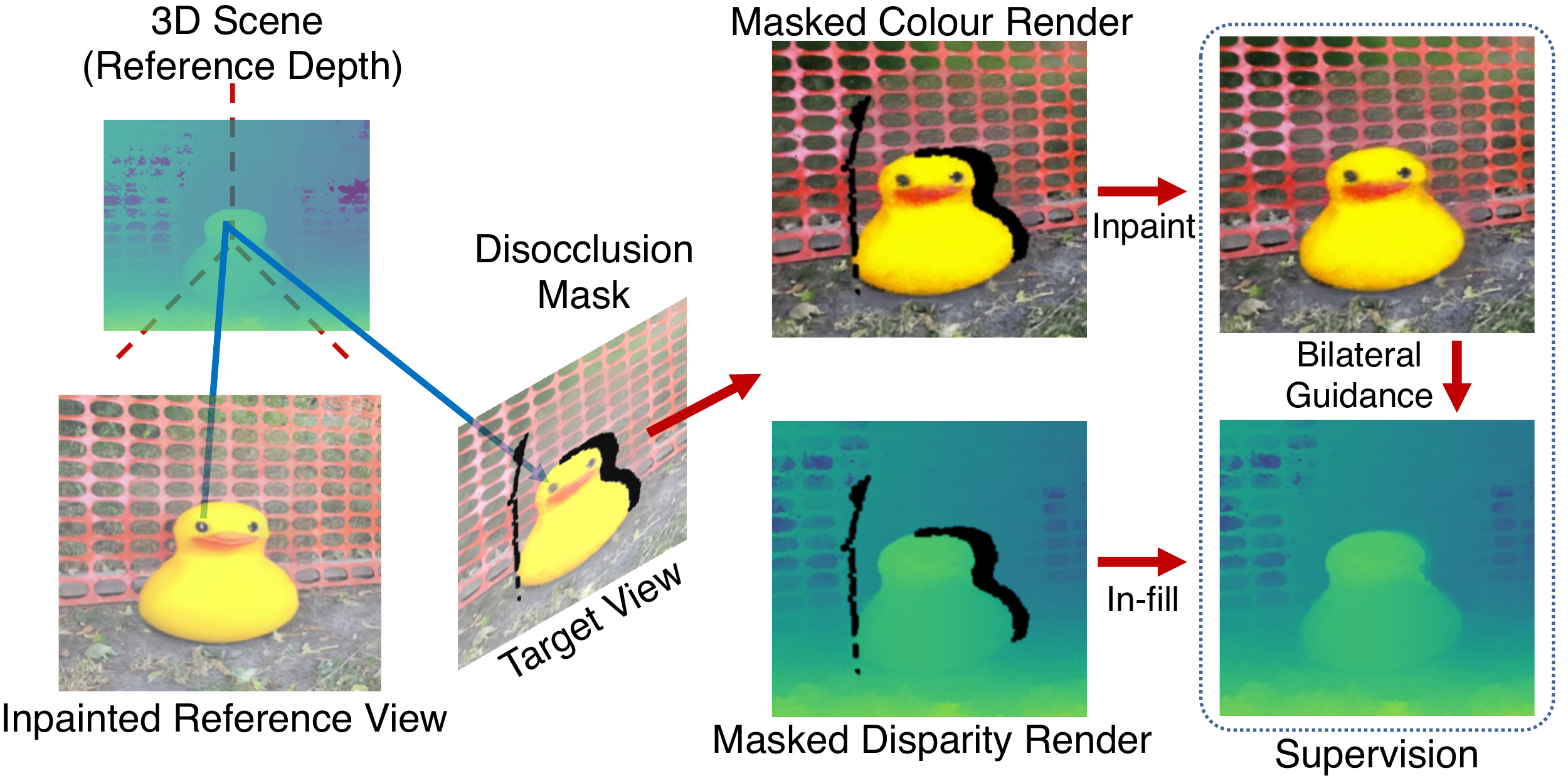}
   \caption{ 
      Overview of our disocclusion handling approach.
      We first identify pixels in the target view, $\Pi_t$, that are \textit{not} visible from the reference view, to build a disocclusion mask, $\Gamma_t$. 
      From $\Pi_t$, 
        we then inpaint a $\Gamma_t$-masked colour render,
        followed by in-filling a disparity render, 
        using bilateral guidance to ensure consistency.
      Finally, these inpainted disoccluded values are used for supervision. %
   }
   \label{fig:disocc}
\end{figure}

While single-reference inpainting prevents problems incurred by view-inconsistent inpaintings, it is missing multiview information in the inpainted region. 
For example, when inserting a duck into the scene, viewing the scene from another perspective naturally unveils new details on and around the duck, due to \textit{disocclusions} 
(see Fig.\ \ref{fig:disocc}). 
We provide an approach to construct these missing details.

Given the inpainted posed reference view, ($I_r$, $\Pi_r$), and a target image, ($I_t$, $\Pi_t$), we first identify the disoccluded pixels in $I_t$ within the mask $M_t$.  
Given the reference disparity image, $D_r$, we unproject every pixel, $p_i\in I_r$, 
into the 3D scene, and then
reproject it into $I_t$, with pixel location $p_{t,i}$.
Every masked pixel in $I_t$ that does \textit{not} receive a projected point (i.e., $\widetilde{p} \in I_t$ s.t.\ $\widetilde{p} \notin \{p_{t,i}\}_i$) is disoccluded; i.e., there is no corresponding pixel in $I_r$ to provide appearance information.
We therefore obtain a disocclusion mask, $\Gamma_t$, associated to $I_t$. 
Next, we inpaint the NeRF render associated to $\Pi_t$, denoted $\widehat{I}_t$, masked by $\Gamma_t$: $\widehat{I}_t^{\,(o)} = \mathrm{Inp}(\widehat{I}_t, \Gamma_t)$. 
Finally, we render the disparity image, $\widehat{D}_t$, and in-fill it as well: 
$ \widehat{D}_t^{(o)} = \mathcal{B}(\widehat{I}_t^{\,(o)},\widehat{D}_t,\Gamma_t) $,
where the bilateral solver, $\mathcal{B}$, 
is guided by the affinities from $\widehat{I}_t^{\,(o)}$ and confidences from $\Gamma_t$.
Similar to \autoref{sec:depth.alignment} and \autoref{sec:view.sub.full}, we recompute $ \widehat{I}_t^{\,(o)} $ and $ \widehat{D}_t^{(o)} $ every $N_\text{do}$ iterations.
For fitting, we use the set of rays from $\Pi_t$ through disoccluded pixels in $I_t$, denoted $\mathcal{R}_{\mathrm{do},t}$ 
(i.e., $r\in \mathcal{R}_{\mathrm{do},t}$ is masked by $\Gamma_t$). 
Over a set of cameras, $T$, the following loss is then used: 
\begin{equation}
\label{eq:disocc}
  \mathcal{L}_{\mathrm{do}} = \mathbb{E}_{t\sim T, r\sim \mathcal{R}_{\mathrm{do},t}}\left[ ||\widehat{C}(r) - C_t(r) ||^2 + \varepsilon(r) \right],
\end{equation}
where 
$\varepsilon(r) = \eta_{\mathrm{do}} [\widehat{D}(r) - D_t(r)]^2$,
$\eta_{\mathrm{do}} > 0$,
and
colour and disparity are $ C_t(r) = \widehat{I}_t^{\,(o)}[r] $ and $ D_t(r) = \widehat{D}_t^{(o)}[r] $.

\section{Experiments}
\label{sec:experiments}

\textbf{Datasets.} 
Following SPIn-NeRF~\cite{spinnerf}, we focus on forward-facing scenes, 
as they are more challenging for the inpainting task.  
For quantitative evaluations, we use the SPIn-NeRF~\cite{spinnerf} dataset, which was designed specifically for 3D inpainting. It contains 10 scenes, each with 60 training views (with the object to be removed), 40 test views (without the object), and human-annotated object masks per view.
For qualitative examples, we adopt forward-facing LLFF scenes~\cite{llff,ibrnet} and the SPIn-NeRF dataset~\cite{spinnerf}. 
We use simple per-scene text prompts to generate inpainted reference views using Stable Diffusion Inpainting v2~\cite{stable.diffusion}; see our supplementary material for details. 

\textbf{Metrics.} 
Given the ill-posed nature of the task,
we follow the 2D~\cite{lama} and 3D~\cite{spinnerf} inpainting literature by evaluating the perceptual quality and realism of the inpainted scenes.
We conduct experiments based on two types of metrics: full-reference (FR) and no-reference (NR). For FR, we compare the inpainted renderings to ground-truth (GT) captures of the scenes \textit{without} unwanted objects, based on LPIPS~\cite{perceptual} and Frechet Inception Distance (FID)~\cite{fid}. For both LPIPS and FID, we only compare the inside of the object bounding boxes, matching SPIn-NeRF's~\cite{spinnerf} setup. For NR, we assess image quality, without using GT captures, by measuring sharpness via the Laplacian variance~\cite{pertuz2013analysis} and MUSIQ~\cite{ke2021musiq}, which uses a learned model of visual quality; see our supplementary material for details. 

\textbf{Baselines.} 
We benchmark our approach against six 3D inpainting models.
(i) \textit{NeRF + LaMa (2D)}: a NeRF is fit to the scene (including the target object), followed by rendering and inpainting via LaMa~\cite{lama} from the test views. %
(ii) \textit{Object-NeRF}~\cite{yang2021learning} directly removes masked points \textit{in 3D}, but does not leverage inpainters to clean up disoccluded regions.
(iii) \textit{Masked NeRF} simply ignores the masked pixels during fitting, relying on the NeRF model itself to interpolate the missing values.
(iv) \textit{NeRF-In}~\cite{nerf.in} uses 2D inpainters as well, including on depth images, but relies on a pixelwise error for fitting, despite multiview inconsistencies.
Two versions of (iv), using \textit{single} and \textit{multiple} inpainted references, are evaluated.
(v) \textit{SPIn-NeRF} \cite{spinnerf} uses a perceptual loss to account for view inconsistencies. We consider two versions with different 2D inpainters, namely Stable Diffusion (SD)~\cite{stable.diffusion} and LaMa~\cite{lama}.
(vi) We also consider a variant of Masked NeRF, with an additional loss based on the recent \textit{DreamFusion}~\cite{poole2022dreamfusion,stable-dreamfusion} model, which utilizes the SD likelihood as a prior for generating textured 3D models. For i, ii, iii, iv, and v we show the results reported in~\cite{spinnerf}. 
See supplementary material for further details.

\begin{table}[tb]
\centering
\caption{
Quantitative full-reference (FR) evaluation of 3D inpainting techniques on the inpainted areas of held-out views from the SPIn-NeRF dataset~\cite{spinnerf}. 
Columns show distance from known ground-truth images of the scene (without the target object), 
    based on a perceptual metric (LPIPS) and feature-based statistical distance (FID).
Our approach with stable diffusion (SD) performs best by both metrics. 
}
\begin{tabular}{lcc}
\hline
\multicolumn{1}{l}{\textbf{Method}} & \textbf{LPIPS}$\downarrow$ & \textbf{FID}$\downarrow$\\ \hline
NeRF + LaMa (2D)~\cite{lama}                                           & 0.5369                         & 174.61 \\
Object NeRF~\cite{yang2021learning}                             & 0.6829                         & 271.80\\
$\mathcal{L}_\text{rec}^\text{unmasked}$ (Masked NeRF)~\cite{original.nerf}                                & 0.6030                         & 294.69\\
$\mathcal{L}_\text{rec}^\text{unmasked}$ + DreamFusion~\cite{poole2022dreamfusion}                        & 0.5934                         & 264.71\\
NeRF-In (multiple)~\cite{nerf.in}                                          & 0.5699                         & 238.33\\
NeRF-In (single)~\cite{nerf.in}                             & 0.4884                         & 183.23\\
SPIn-NeRF-SD~\cite{spinnerf}                                    & 0.5701                         & 186.48\\
SPIn-NeRF-LaMa~\cite{spinnerf}                                  & 0.4654                         & 156.64\\ \hdashline
Ours-SD                                                         & \textbf{0.4532}                & \textbf{116.24}\\ \hline
\end{tabular}
\label{tab:multiview.inpainting}
\end{table}

\begin{table}[tb]
\centering
\caption{
Quantitative no-reference (NR) evaluation of 3D inpainting on videos rendered from the SPIn-NeRF dataset. 
Our approach outperforms SPIn-NeRF (the second-highest performing model according to the full-reference metrics). %
}
\begin{tabular}{lcc}
\hline
\multicolumn{1}{l}{\textbf{Method}} 
    & \textbf{Sharpness}$\uparrow$  
    & \textbf{MUSIQ}$\uparrow$ \\ \hline
SPIn-NeRF-LaMa~\cite{spinnerf}                          & 354.31 & 58.10 \\
Ours-LaMa                                               & 394.55 & 62.00 \\ 
Ours-SD                                                 & 398.56 & 61.47 \\ \hline
\end{tabular}
\label{tab:multiview.inpainting.sharpness}
\end{table}

\begin{figure}[t]
  \centering
   \includegraphics[width=0.99\linewidth]{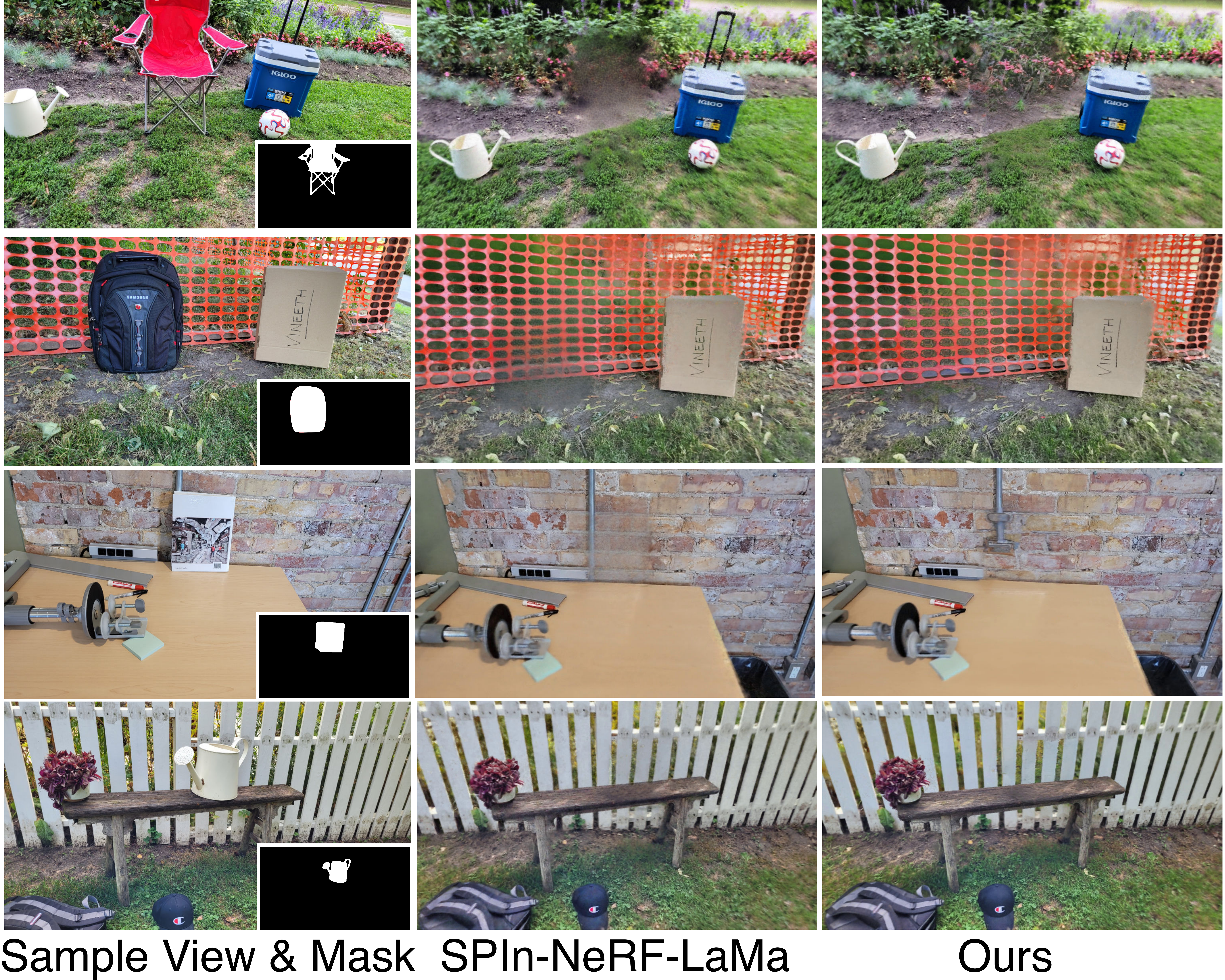}
   \caption{ 
   Qualitative comparison of novel view renderings of our method with SPIn-NeRF-Lama (the second-best model quantitatively). 
   We find that SPIn-NeRF still outputs blurry textures in the masked area (see first three rows), 
   while ours is always sharp.
   See our supplementary material for additional comparisons with other baselines. 
   }
   \label{fig:spinnerf.nerfin.compare}
\end{figure}

\begin{figure}[t]
  \centering
   \includegraphics[width=0.99\linewidth]{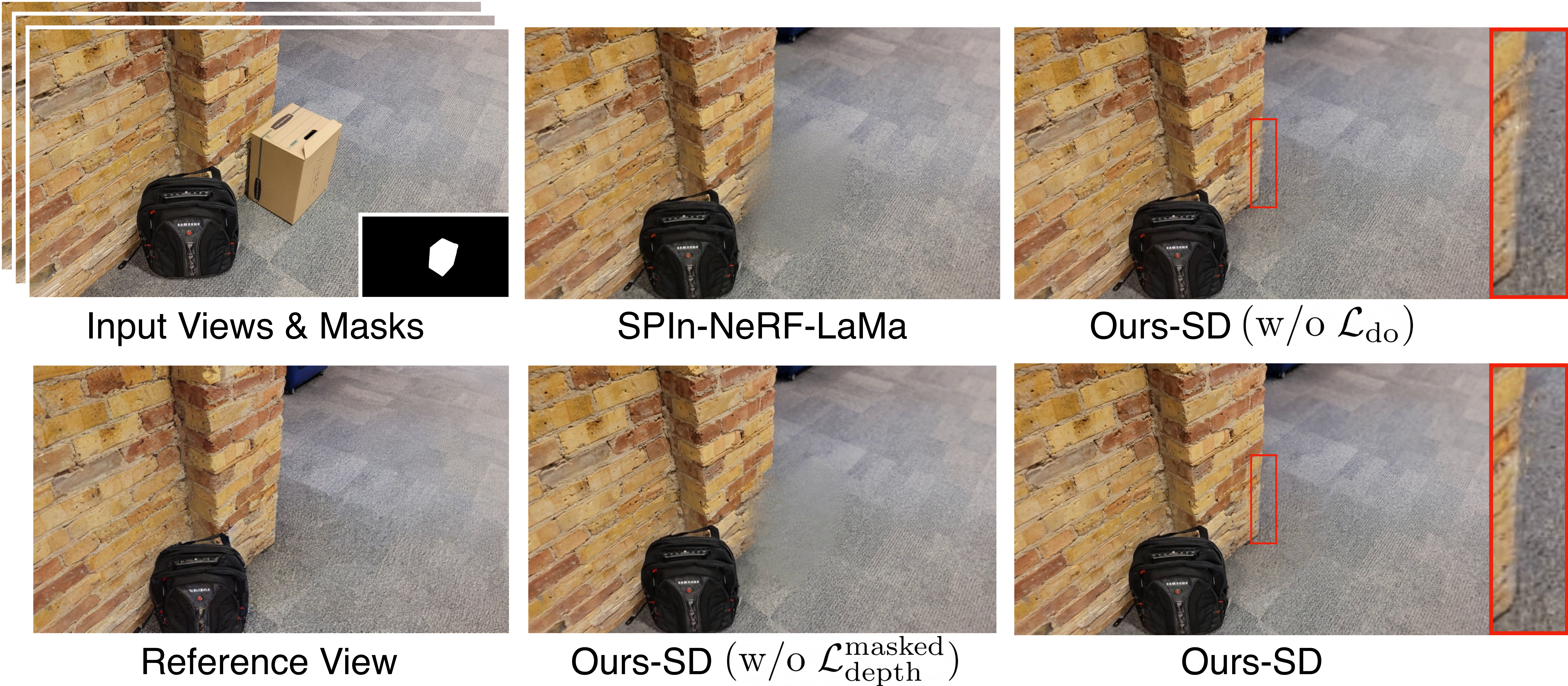}
   \caption{ 
   Qualitative example of effects of ablation.
    Removing the masked depth loss catastrophically damages the geometric structure of the inpainted area
    (lower-centre inset).
    The upper-right image shows that
        ablating the disocclusion loss results in a blurrier output near the edge of the wall (see zoomed-in area),
    while the inset at the bottom-right (the full model)
        has sharper novel structure. 
   }
   \label{fig:ablation}
\end{figure}

\begin{figure}[t]
  \centering
   \includegraphics[width=0.99\linewidth]{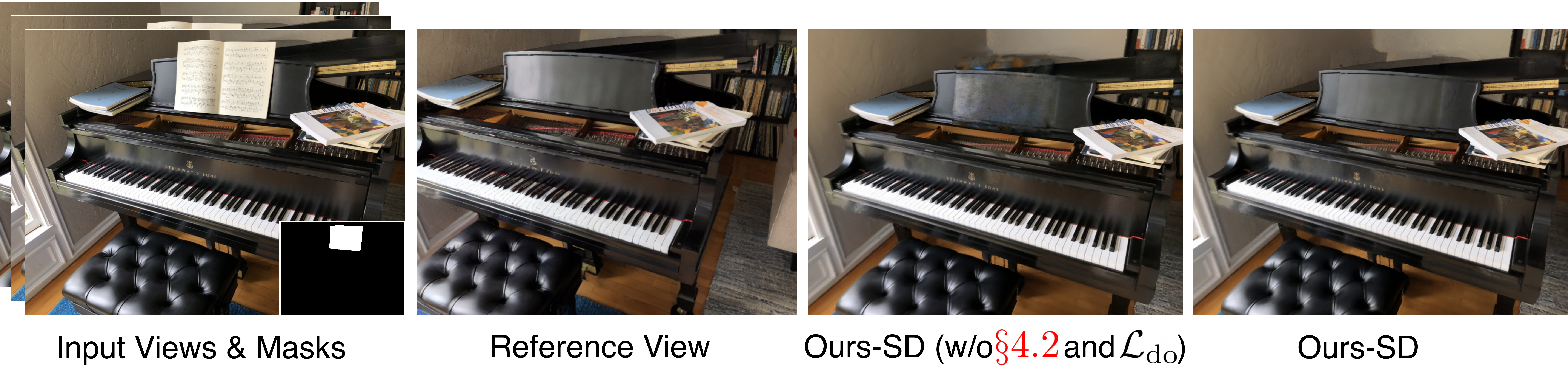}
   \caption{ 
        Visualization of ablated view-dependent-effects handling.
        Visual quality degrades in the ablated scenario 
        (without \autoref{sec:view.sub.full} and $\mathcal{L}_\text{do}$ from \autoref{sec:disoccluded.regions}), 
            with rough, uneven lighting across the masked area
            and 
            an unrealistic jump in brightness on the left mask edge.
        In contrast, the full model (rightmost inset) smoothly interpolates the view-dependent lighting of the novel view into the masked area. 
   }
   \label{fig:view.dependence.ablation}
\end{figure}

\begin{figure}[t]
  \centering
   \includegraphics[width=0.99\linewidth]{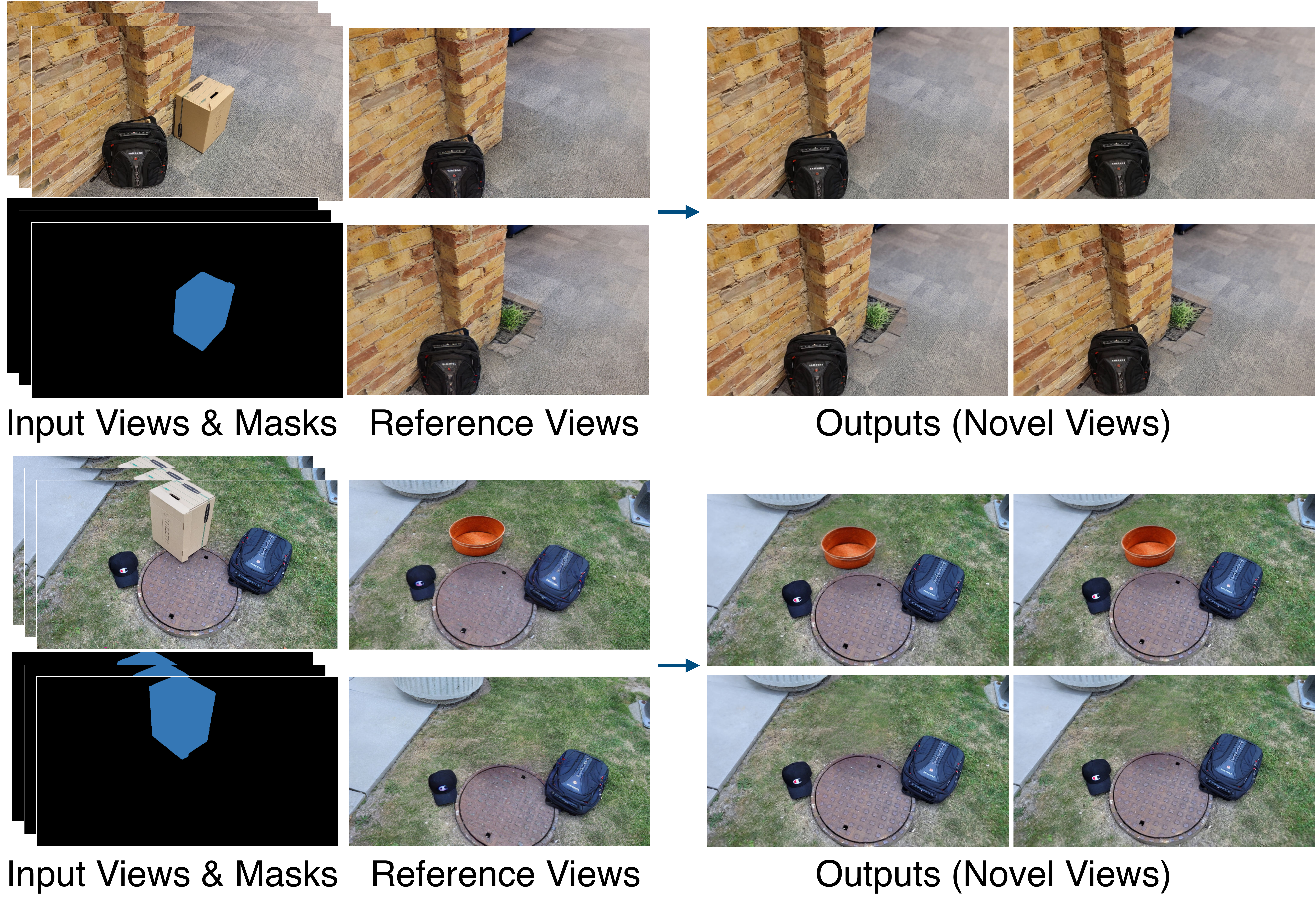}
   \caption{ 
   Qualitative illustration of our results on two scenes from the SPIn-NeRF dataset~\cite{spinnerf}. For each scene, we use two different reference views to generate corresponding inpainted scenes. For each inpainted scene, we show two novel view renderings. 
   Note the ability to insert novel content into the 3D scene.
   Please see our supplementary material and \href{https://ashmrz.github.io/reference-guided-3d}{website} for additional visualizations. 
   }
   \label{fig:controllable}
\end{figure}

\textbf{Quantitative Results.}
In Table~\ref{tab:multiview.inpainting}, 
    we see that our approach provides the best performance on both FR metrics.
The Object-NeRF and Masked-NeRF approaches, which perform object removal without altering the newly revealed areas, perform the worst.
Combining Masked-NeRF with DreamFusion performs slightly better. This indicates some utility of the diffusion prior;
however, while DreamFusion can generate impressive 3D entities in isolation, it does not produce sufficiently realistic outputs for inpainting real scenes. 
SPIn-NeRF-SD obtains a similar poor LPIPS, though with better FID. It is unable to cope with the greater mismatches of the SD generations.
NeRF-In outperforms the aforementioned models. Still, the use of a pixelwise loss leads to blurry outputs. %
Finally, our model outperforms the second-best model (SPIn-NeRF-LaMa) considerably in terms of FID, reducing it by ${\sim}$25\%.

FR measures are limited by their use of a single GT target image. 
We therefore also examine NR performance, 
demonstrating
improvements over SPIn-NeRF, 
    in terms of both sharpness (by $11.2$\%) 
    and MUSIQ (by $5.8$\%); see Table~\ref{tab:multiview.inpainting.sharpness}.
This confirms our qualitative observation (see Fig.~\ref{fig:spinnerf.nerfin.compare}) that our results are considerably sharper and more realistic.

\textbf{Ablations.}
In Table~\ref{tab:multiview.inpainting.ablation}, we illustrate the effect of ablating components of our algorithm. 
Using LaMa to obtain $I_r$ led to inferior performance, showing our model benefits from better image inpainters. 
Further, 
each of the VDE (\autoref{sec:view.sub.full}), 
masked depth ($\mathcal{L}_\text{depth}^\text{masked}$ from Eq.~\ref{eq:aligned.depth} in \autoref{sec:depth.alignment}), and disocclusion ($\mathcal{L}_\text{do}$ from Eq.~\ref{eq:disocc} in \autoref{sec:disoccluded.regions}) 
handling
improve 
quality. 
We also include a ``gold standard scenario'', where a real photo is used instead of an inpainted one, loosely indicating the best possible score one can expect from the model;
this suggests there is room for improvement, simply by improving inpainted reference views.
Qualitatively, Fig.~\ref{fig:ablation} illustrates the results of ablating 
$\mathcal{L}_\text{depth}^\text{masked}$ and $\mathcal{L}_\text{do}$. The former is harmful to geometric quality (and thus image structure) while the latter blurs outputs in disoccluded areas. 
Another core contribution is our ability to handle 
    VDEs
    in non-reference views;
    ablating our view-substitution-based technique %
    degrades
    visual quality, 
        as shown in Fig.~\ref{fig:view.dependence.ablation},
        with uneven and unrealistic brightnesses in novel views.
Our supplement contains additional visualizations. %

\begin{table}[tb]
\centering
\caption{
Quantitative evaluation of methodological ablations via full-reference (FR) metrics.
Removing the contribution of VDEs via view-substitution,
masked depth loss,
and disocclusion handling
lead to reduced FR performance.
To demonstrate the potential of improving reference image quality, we evaluate a model using a GT capture as $I_r$. 
}
\begin{tabular}{lcc}
\hline
\multicolumn{1}{l}{\textbf{Method}} & \textbf{LPIPS}$\downarrow$ & \textbf{FID}$\downarrow$\\ \hline
Ours-LaMa                                                       & 0.4634                         & 133.27\\
Ours-SD (w/o~\autoref{sec:view.sub.full} and $\mathcal{L}_\text{do}$)      
                                                                & 0.5279                         & 145.60\\
Ours-SD (w/o $\mathcal{L}_\text{depth}^\text{masked}$)          & 0.5211                         & 181.20\\
Ours-SD (w/o $\mathcal{L}_\text{do}$)                           & 0.4676                         & 126.74\\ 
Ours-SD                                                         & 0.4532               & 116.24\\ \hdashline
Ours (w/ GT reference view)                                     & 0.3889                         & 104.10\\ \hline
\end{tabular}
\label{tab:multiview.inpainting.ablation}
\end{table}

\textbf{Controllability.}
An additional major capability of our method is the ability to insert novel content into the 3D scene by providing a different inpainted single-image reference. We refer to this as \textit{controllability} and showcase examples in Fig.~\ref{fig:overall}.
While other methods can also insert content by altering one view,
    such as using NeRF-In with a single reference, 
ours 
    (i) avoids visual quality degradation in views far from the reference and
    (ii) generates non-reference VDEs as well.
We demonstrate this in Fig.~\ref{fig:controllable},
where we add novel content to each scene, %
    such as the indoor garden and wash basin (see supplemental for more examples).
We remark that the expanding generative capacity and creativity of 2D inpainting models, 
such as text-guided diffusion models 
(e.g., \cite{stable.diffusion,ramesh2022hierarchical}),
will render controllability increasingly important in future work. %

\section{Conclusion}
In this paper, we presented an approach to inpaint NeRFs, via a single inpainted reference image. 
We used a monocular depth estimator, aligning its output to the coordinate system of the inpainted NeRF to back-project the inpainted material from the reference view into 3D space. 
We further leveraged bilateral solvers to add VDEs to the inpainted region, and used 2D inpainters to fill disoccluded areas. 
Our work still has two main limitations: first, we fall back to a diffuse prior in the case of masked edge islands (i.e., when we cannot hallucinate VDEs). Second, exact depth alignment remains difficult.
Still, using  multiple evaluation metrics, we demonstrated the superiority of our algorithm over prior 3D inpainting methods. %
We also illustrated the controllability advantage of our model, 
enabling users to easily alter the generated scene through a single guidance image.

{\small
\bibliographystyle{ieee_fullname}
\bibliography{egbib}
}

\newpage
\clearpage

\appendix

\section{Summary}
We provide additional details about the comparative baselines, against which we benchmark, in~\autoref{sec:app:baselines}.
Further exposition about our disparity smoothing technique (see~\autoref{sec:depth.alignment}) and edge island filtering method (see~\autoref{sec:supervision.from.ref}) is given in~\autoref{app:sec:disp.smoothing} and \autoref{sec:edgeislands}, respectively.
Additional visualizations are shown in~\autoref{app:sec:additional.vis}, including methodological illustrations~(\autoref{app:sec:addvis.method}) and qualitative examples~(\autoref{app:sec:addvis.ablations} and \autoref{app:sec:addvis.qual}).
Technical implementation details, such as hyper-parameter values, are discussed in~\autoref{sec:implementation.details}.
Finally, further explanation about our choice of evaluation metrics is given in~\autoref{app:sec:metrics}.
Please also view our supplementary website for additional visualizations, including videos.

\section{Baseline Details} \label{sec:app:baselines}

\subsection{Masked-NeRF + DreamFusion}

For the \textit{Masked-NeRF + DreamFusion} baseline, we use the same per-scene text prompts we used to generate our reference views, to guide the generation of the masked region using the score distillation sampling (SDS)~\cite{poole2022dreamfusion} loss. We found that gradually and uniformly decreasing the maximum noise steps, $t_\text{max}$, during fitting, until it equals the minimum noise steps, $t_\text{min}$, at the last iteration, improves quality. We suggest this is because, at first, higher noise levels are effective in the generation of global scene structure, and later, lower noise-levels enable fixing details. 
Due to the unavailability of DreamFusion's code and their underlying diffusion model, Imagen~\cite{imagen}, we used stable-dreamfusion~\cite{stable-dreamfusion}, with Stable-Diffusion~\cite{stable.diffusion} as the underlying diffusion model.

\subsection{NeRF-In}
As in prior work~\cite{spinnerf}, we used our own implementation of NeRF-In~\cite{nerf.in}, due to the unavailability of official code. 
Besides the primary distinctions with our method, such as the pixelwise loss, the remaining architecture (e.g., the use of NGP~\cite{instant.ngp}) is identical to our method.
Note that this induces minor implementation differences from the concurrent technical report of NeRF-In, such as the choice of pretrained 2D inpainting model.

Since NeRF-In considers the effect of varying numbers of reference images, we considered two variants of NeRF-In: using multiple reference images (i.e., inpainting all images, as in SPIn-NeRF~\cite{spinnerf} and using a single one.
By default, we utilize the latter method, as it obtains better overall performance (in both our experiments and those of NeRF-In itself), but report the performance of both models in Table~\ref{tab:multiview.inpainting}. 

\subsection{Object-NeRF}
Following the Object-NeRF~\cite{yang2021learning} model, 
we can remove objects by simply ignoring the contribution of masked 3D points in the volume rendering process (equivalent to setting $\sigma_i = 0$ in masked regions).
This is possible here due to the assumed availability of a 3D mask.
Note that we are only utilizing this particular approach to object removal, not the entire Object-NeRF algorithm (i.e., the construction of the NeRF itself is identical to our method).

\section{Disparity Smoothing Details}
\label{app:sec:disp.smoothing}
After performing the initial depth alignment (as discussed in \autoref{sec:depth.alignment}), 
we further reduce the misalignments around the edges of the reference mask, $M_r$, via smoothing the aligned reference disparity, $D_r$.
More specifically, to improve the visual continuity of the reference-view boundary between the aligned masked disparity, $D_r \odot M_r$, and the unmasked rendered NeRF disparity, $\hat D_r \odot (1 - M_r)$, we smooth $D_r$ to get the edge-smoothed disparity, $D_r^\text{smooth}$:
\begin{equation}
    \label{eq:disparity.smoothing}
    D_r^\text{smooth} = D_r + D^\text{correction},
\end{equation}
where $D^\text{correction}$ is the smoothed disparity correction obtained by minimizing the following objective:
\begin{align}
    \label{eq:disparity.error.smooth}
    &\big\Vert (\hat D_r - D_r^\text{smooth}) \odot (1 - M_r) \big\Vert_2^2 \notag \\
    +&\gamma_\text{smooth} \sum_{p \in I_r} \sum_{p' \in \mathcal{N}(p)} \big(D^\text{correction}(p) - D^\text{correction}(p')\big)^2,
\end{align}
where for a pixel, $p$, $\mathcal{N}(p)$ is the set of four neighbouring pixels, and $\gamma_\text{smooth}$ is the weight of the smoothness loss. The first term in Eq.~\ref{eq:disparity.error.smooth} fits the unmasked pixels of $D^\text{correction}$ to the difference of the rendered disparity, $\hat D_r$, and the aligned disparity, $D_r$. The second term is the smoothness penalty, to smoothly propagate the values of $D^\text{correction}$ from outside the mask to inside.

\begin{figure}[t]
  \centering
   \includegraphics[width=0.99\linewidth]{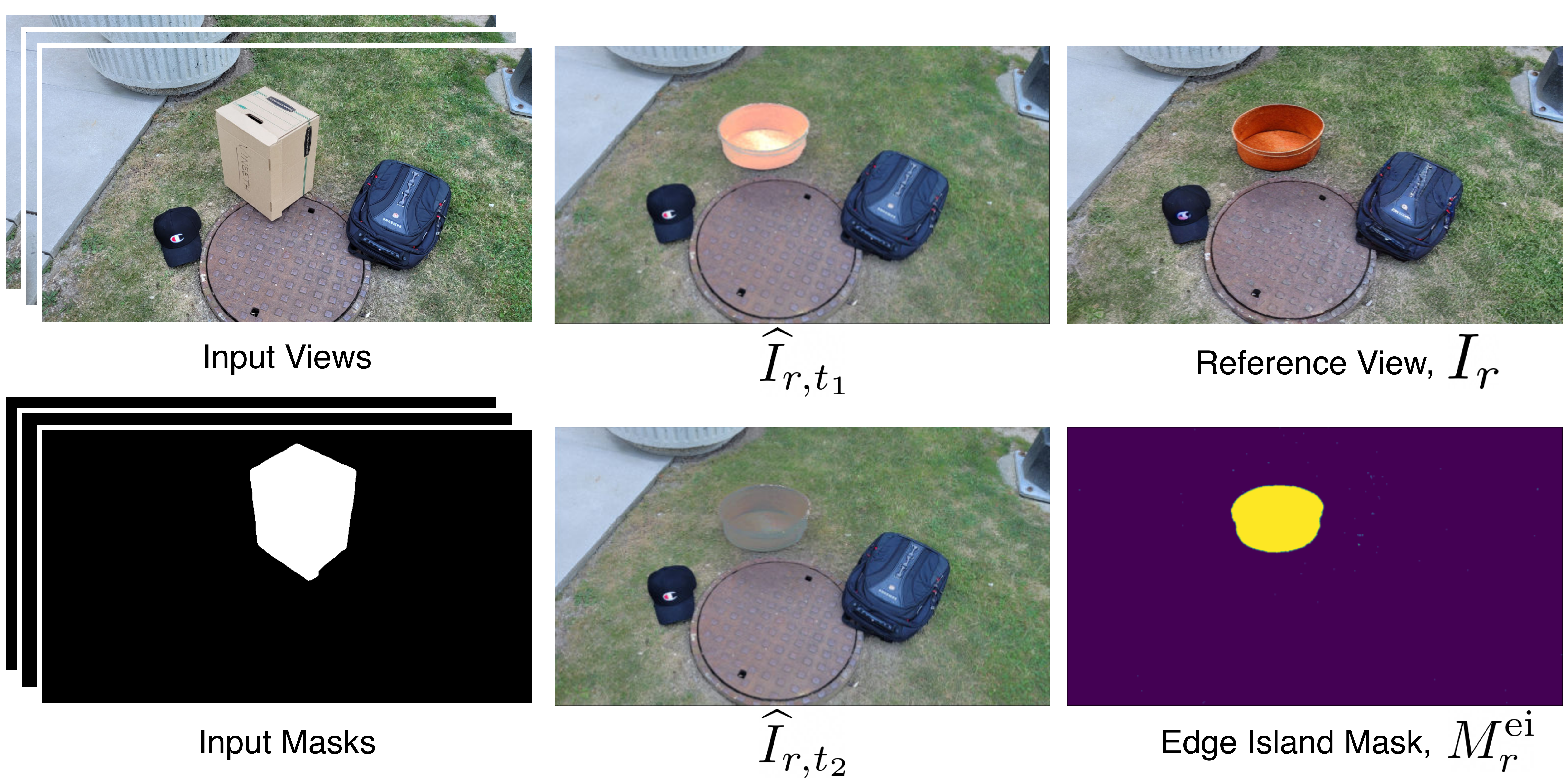}
   \caption{ 
        Examples of our ``edge island'' detection method, designed to filter out erroneous outputs from the bilateral filter (detailed in \autoref{sec:edgeislands}).
        Left column: input views and masks for the scene.
        Middle column: view-substituted renders after bilateral inpainting (see also \autoref{sec:view.sub.full}), which has produced poor quality colours in the edge island formed by the washtub.
        Right column: (top) the reference view and (bottom) the detected mask, used to filter out rays that would potentially damage the output.
   }
   \label{fig:edge.island.filtering}
\end{figure}

\begin{figure}[tb]
  \centering
   \includegraphics[width=1.0\linewidth]{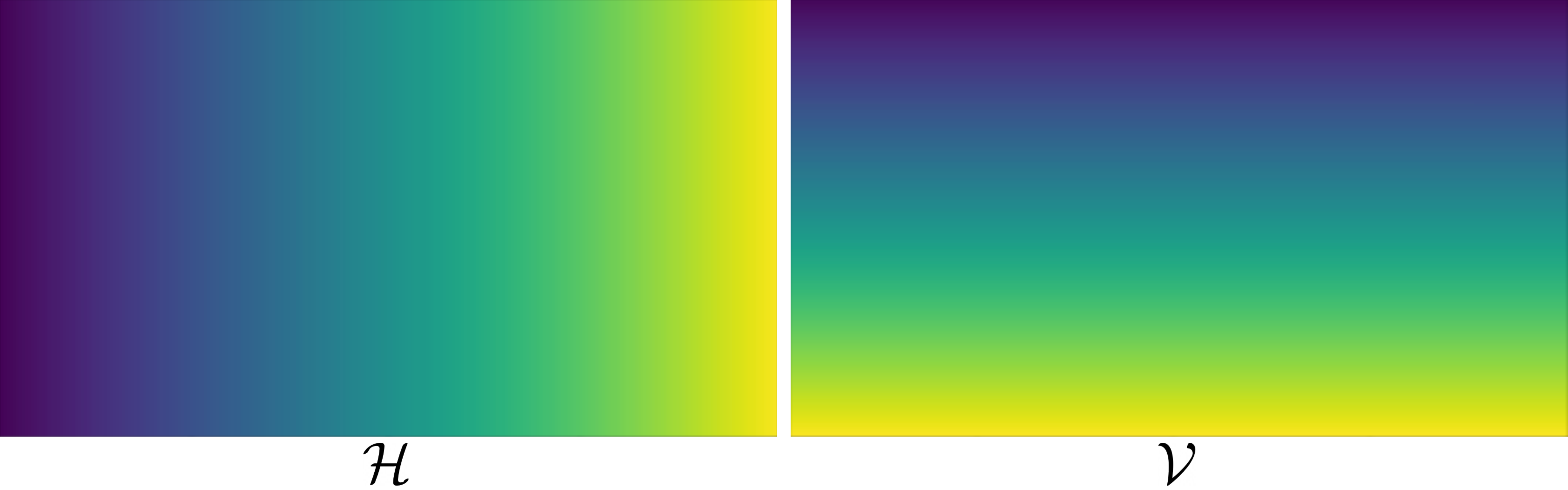}
   \caption{The additional matrices used for tighter alignment around the edges when aligning disparities (see \autoref{sec:depth.alignment}). In our experiments, scale and offset were insufficient to have the depths completely aligned around the boundaries of the mask. These two matrices allow the predicted depth to be tilted along the $x$ and $y$ axes. %
   }
   \label{app:fig:kernels}
\end{figure}

\section{Edge Island Filtering Details} \label{sec:edgeislands}

When propagating appearance information into the masked area, in order to construct view-dependent effects for supervision in non-reference views, recall that the bilateral solver is sometimes unable to provide sensible colour values in some areas of the masked region, due to the presence of ``edge islands'' (see \autoref{sec:supervision.from.ref}).
Such areas are isolated patches in bilateral space, for which the bilateral solver cannot effectively produce colour values (see Fig.~\ref{fig:edge.island.filtering} for instances of this).
In this section, we provide additional details on our filtering algorithm for removing these invalid  values, so that they are not used for supervision. 

First, we dilate the mask, $M_r$, with kernel size $5$ to get the dilated mask, $M_r^\text{dilated}$. Then, for each target view, $t$, we find the maximum absolute value of the residual inside $M_r^\text{dilated}$ and outside $M_r$:
\begin{equation}
    \text{res}_t^{\max} = \max \Big( \text{abs} (\text{res}_t) \odot \big(M_r^\text{dilated} \cap (1 - M_r) \big) \Big),
\end{equation}
where $\text{abs}(\cdot)$ is the element-wise absolute value. 
We denote the mask for the pixels in $res_t \odot M_r$ with absolute values higher than $\text{res}_t^{\max} \times c_\text{ei}$ as $M_{r,t}^{ei}$, where $c_\text{ei} \geq 1$ is the filtering threshold. The mask of the edge island is then obtained as the union of the mask of all of the out-of-distribution values among all of the target views:
\begin{equation}
    M_r^\text{ei} = \bigcup_t M_{r,t}^{ei}.
\end{equation}
Fig.~\ref{fig:edge.island.filtering} shows an example of the effects of an edge island inside the masked region (the orange pan) on the target colours
of two example target views,
$\widehat I_{r, t_1}$ and $\widehat I_{r, t_2}$. 
As shown in the figure, the bilateral solver has failed to predict correct view-dependent colours for the pan, resulting in extreme behaviour inside the pan. 
Our proposed edge island filtering successfully  detects and removes the outlier values via the edge island mask, $M_r^\text{ei}$.

\begin{figure*}[ht]
  \centering
   \includegraphics[width=1.0\linewidth]{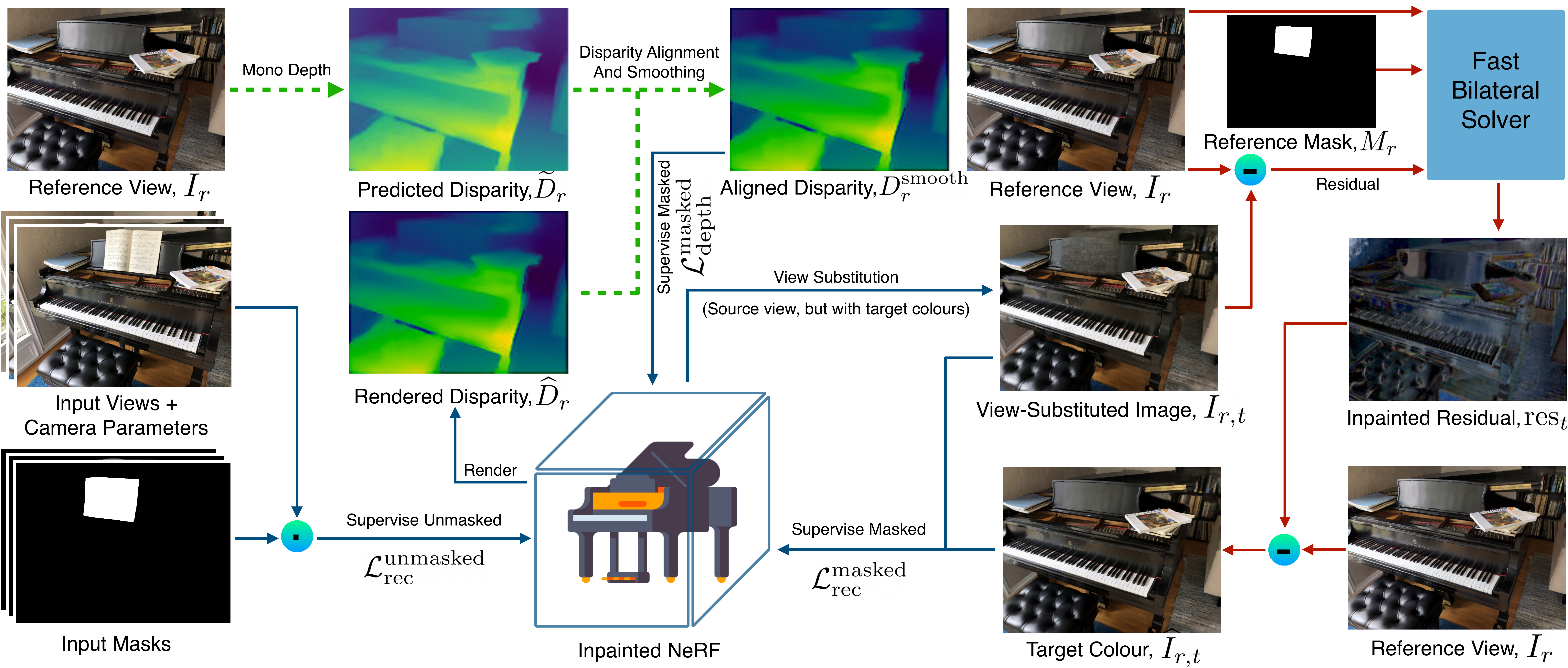}
   \caption{
    Schematic overview of our NeRF fitting algorithm for 3D inpainting.
    The inputs to the method are a single inpainted reference view, $I_r$, 
        and a set of posed images with associated inpainting masks (leftmost column). 
    We begin the fitting process with standard NeRF supervision on the \textit{un}masked areas of the images,
        after which we can render a disparity map, $\widehat{D}_r$, with reasonable quality outside the mask (lower-left insets).
    We then use a monocular depth estimator to obtain the predicted disparity, $\widetilde{D}_r$, and apply a novel alignment procedure (\autoref{sec:depth.alignment}) to obtain an aligned disparity map, $D^{\text{smooth}_r}$, which can be used to supervise the depth \textit{under} the mask via loss $\mathcal{L}_\text{depth}^\text{masked}$ (upper middle inset).
    Finally, to obtain view-dependent effects in unseen views (\autoref{sec:view.sub.full}), 
        we utilize our new \textit{view-substitution} technique (\autoref{sec:view.sub}) to render an image, $I_{r,t}$, via the reference camera, but with the colours of a non-reference (target) view, $I_t$ (centre-right inset).
    The view-substituted image, $I_{r,t}$, is subtracted from the reference view, $I_r$, to obtain a residual image, $\Delta_t = I_r - I_{r,t}$; 
    we then apply the bilateral solver, $\mathcal{B}$, to refine $\Delta_t$, using the reference mask, $M_r$, to construct a  confidence map (low inside the mask and high outside it), guided by the bilateral affinities of $I_r$ (upper-right insets; see \autoref{sec:bilateral.inp.residual}).
    This has the effect of ``diffusing'' the view-dependent effects of the non-reference view from \textit{outside the mask} into the inside of the masked area, obtaining an ``inpainted'' residual, $\text{res}_t$.
    Subtracting this from $I_r$ gives our desired colours, $\widehat{I}_{r,t} = I_r - \text{res}_t$, 
        which can be used to supervise the colours \textit{under} the mask (lower-right insets).
    The resulting combined losses thus supervise the NeRF from non-reference target viewpoints both outside the mask ($\mathcal{L}_\text{rec}^\text{unmasked}$) and inside the mask ($\mathcal{L}_\text{depth}^\text{masked}$ and $\mathcal{L}_\text{rec}^\text{masked}$). %
    See \autoref{sec:method} for details.\protect\footnotemark
   }
   \label{fig:method.long}
\end{figure*}

\footnotetext{IBRNet images in Fig. 13,14 by Wang et al. available in \href{https://drive.google.com/drive/folders/1qfcPffMy8-rmZjbapLAtdrKwg3AV-NJe}{IBRNet}~\cite{ibrnet} 
under a \href{https://creativecommons.org/licenses/by/3.0}{CC BY 3.0 License}.}

\begin{figure*}[t]
  \centering
   \includegraphics[width=0.99\linewidth]{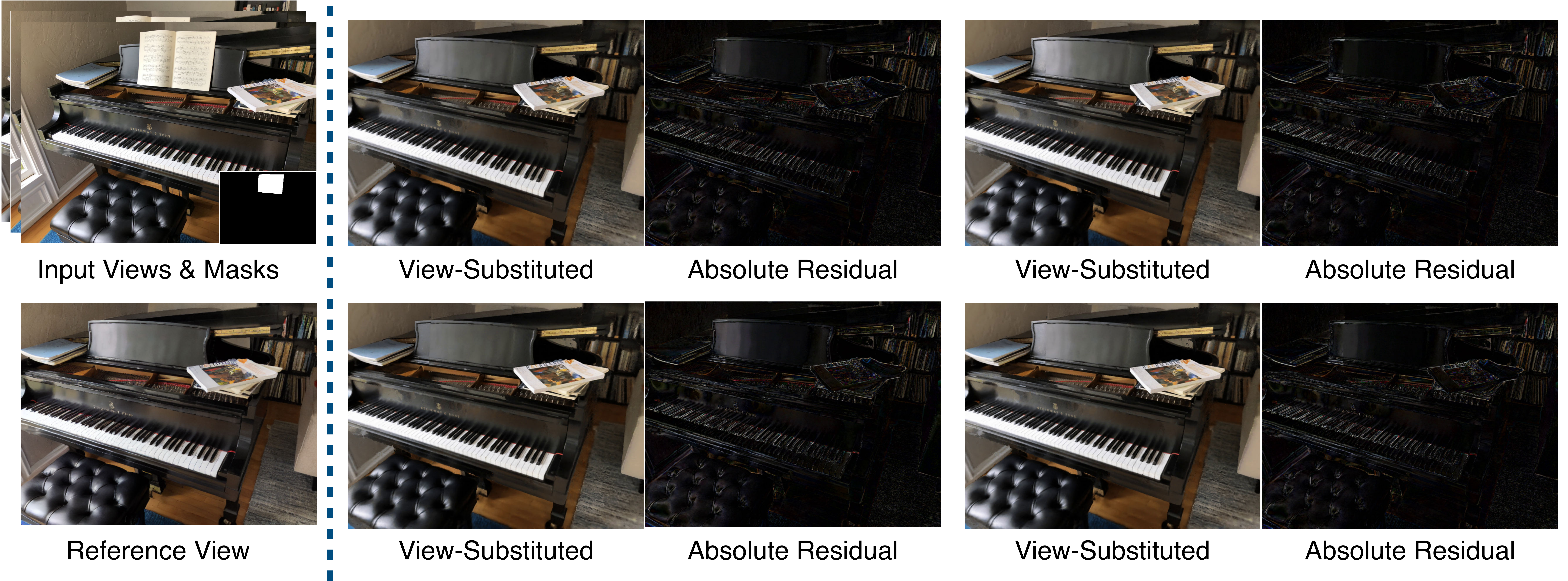}
   \caption{ 
   Overview of the outputs of our view-substitution method. The input views and masks (top-left) with their corresponding camera parameters, in addition to a single reference view (bottom-left), are the inputs to our multiview inpainting approach. On the right hand side, we show the view-substituted renderings, $\{I_{r,t_1}, \cdots, I_{r,t_4}\}$,  for four different target views, $\{t_1, \cdots, t_4\}$. For each view-substituted image, $I_{r,t}$, we also provide the absolute value of the residual, $\vert I_r - I_{r, t} \vert$, to illustrate the view-dependent effects provided by our approach. Notice that all of the view-substituted images are looking at the scene from the reference camera, but the rendered colours are from different target cameras. 
   } 
   \label{fig:view.sub.examples}
\end{figure*}

\begin{figure}[t]
  \centering
   \includegraphics[width=0.99\linewidth]{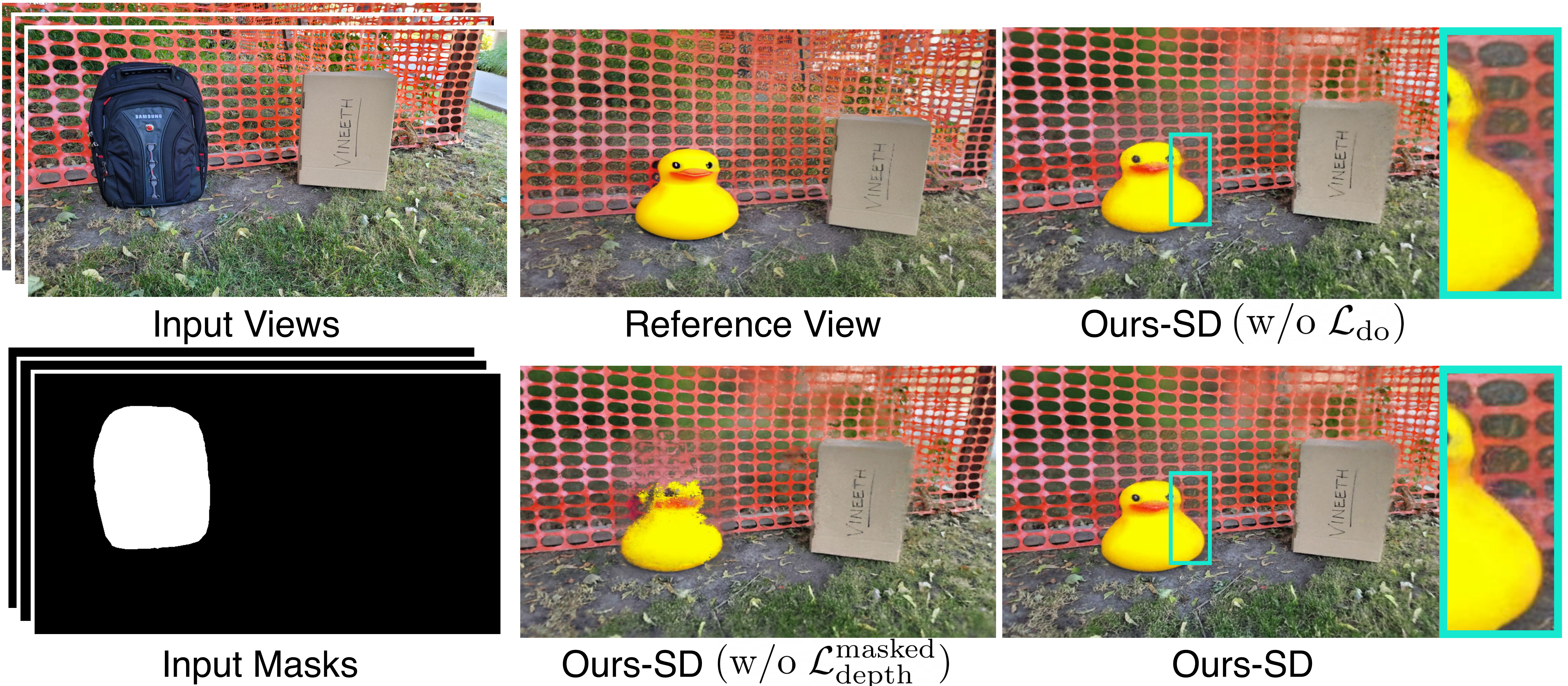}
   \caption{ 
    Qualitative example of effects of ablation (see also Fig.~\ref{fig:ablation}).
    Notice the degradation incurred by not using the masked depth supervision (lower-middle inset) and the slightly blurrier outputs in the disoccluded region when not using $\mathcal{L}_\text{do}$ (upper-right inset; look closely at the zoomed area, particularly at the background close to the edge of the inserted duck).
   }
   \label{fig:ablation.duck}
\end{figure}

\begin{figure}[t]
  \centering
   \includegraphics[width=0.99\linewidth]{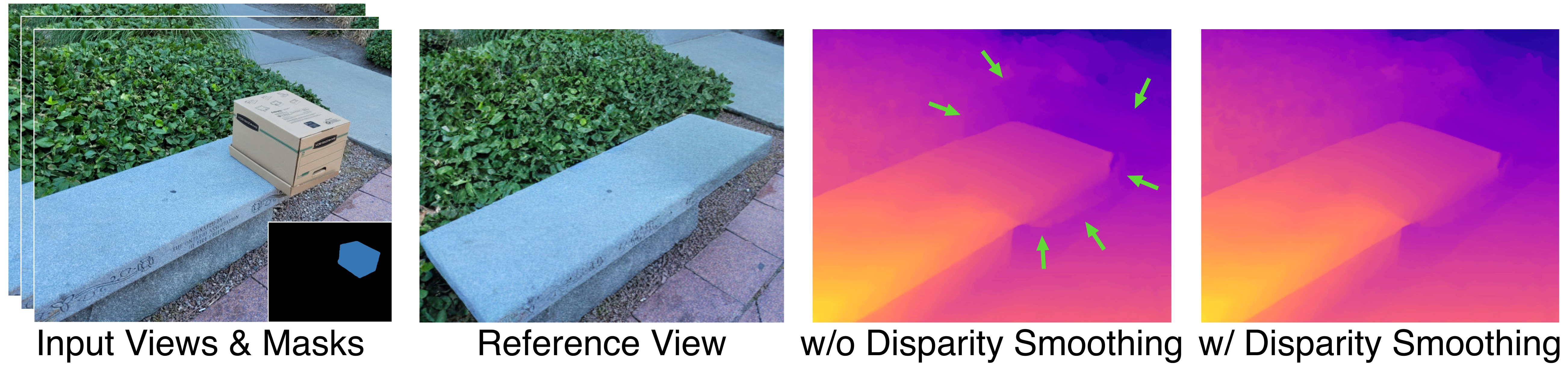}
   \caption{ 
   Effect of our disparity smoothing step (see \autoref{sec:depth.alignment} and \autoref{app:sec:disp.smoothing}) on the rendered disparities. As illustrated above, the edges of the masked region (around the box) are more blended in with the surrounding after adding the disparity smoothing component. 
   }
   \label{fig:smoothing.ablation}
\end{figure}

\begin{figure*}[t]
  \centering
   \includegraphics[width=0.99\linewidth]{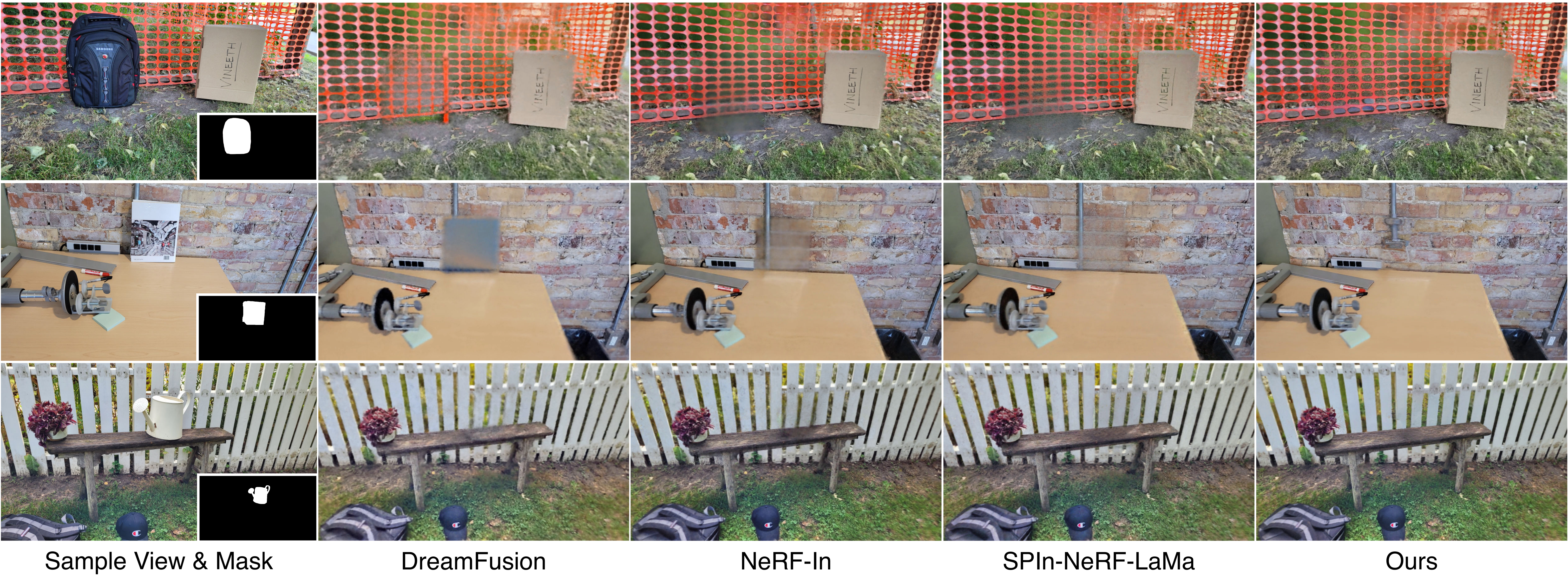}
   \caption{ 
   Additional qualitative comparisons to baselines with synthesized novel views.
   The Masked-NeRF+DreamFusion model (second column) does improve quantitatively (see Table~\ref{tab:multiview.inpainting}) over using Masked-NeRF alone or simply removing the object in 3D without inpainting (the ``Object-NeRF'' baseline), but it does not output sufficiently realistic details to outperform our method: see the oversaturated colours on the fence in first row and the unnatural output in the second row.
   NeRF-In~\cite{nerf.in} (third column), here using the ``multiple'' variant with LaMa~\cite{lama},
   is quite blurry, due to disagreements between inpainting details among the input images.
   SPIn-NeRF~\cite{spinnerf} (fourth row) improves on this via the use of a perceptual loss~\cite{perceptual}, but still generates blurry details when significant disagreement among inpaintings are present (semantic differences, as such the presence or absence of the pipe in the second row, and complex textures (e.g., the grassy dirt in row one or the variously coloured bricks in row two) can exacerbate this problem).
   In contrast, our method is consistently sharp;
      see also Fig.~\ref{fig:spinnerf.nerfin.compare}.
   } 
   \label{fig:baseline.qualitative.nerfin.compare}
\end{figure*}

\begin{figure}[t]
  \centering
\adjincludegraphics[width=0.99\linewidth,trim={{.505\width} 0 0 0},clip]{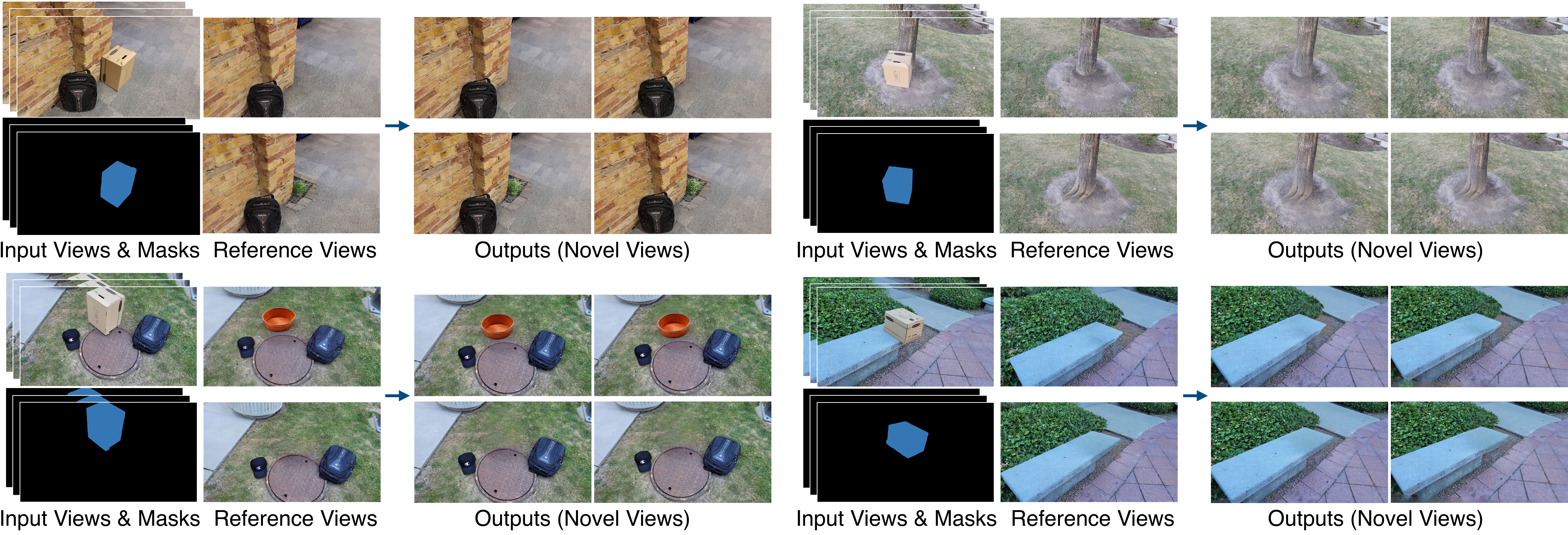}
   \caption{ 
   Qualitative illustration of our results on additional scenes from the SPIn-NeRF dataset~\cite{spinnerf};
   see also Fig.~\ref{fig:controllable}.
   For each scene, we use two different reference views to generate corresponding inpainted scenes. For each inpainted scene, we show two novel view renderings. 
   Note the ability to insert novel content into the 3D scene or modify existing scene structure,
    such as adding the tree roots
    and controlling the length of the bench.
   Please see our supplementary website for additional visualizations. 
   }
\label{fig:controllable.second}
\end{figure}

\section{Additional Visualizations} \label{app:sec:additional.vis}

\subsection{Methodological Illustrations} \label{app:sec:addvis.method}

\textbf{Depth Alignment Tilt Matrices.}
In Fig.~\ref{app:fig:kernels}, we visualize the matrices utilized for tighter depth alignment (see~\autoref{sec:depth.alignment}).
These matrices allow the optimization to \textit{tilt} the depths, in addition to scaling and shifting them.

\textbf{Overview.}
We provide an expanded methodological illustration in Fig.~\ref{fig:method.long}, covering our approach to providing geometric supervision in the masked region (\autoref{sec:depth.alignment}) and handling the construction of view-dependent effects in non-reference views (\autoref{sec:view.sub.full});
see also Figs.~\ref{fig:method}, \ref{fig:view.substitution}, and \ref{fig:view.sub.summary}.

\textbf{View-Substituted Images.}
We also provide some examples of view-substituted images (see \autoref{sec:view.sub}) in Fig.~\ref{fig:view.sub.examples}. 
Notice that the view-substituted images have identical camera viewpoint (and thus image structure) as the reference image, but different colours, corresponding to the view-dependent visual differences across the non-reference images. 

\subsection{Additional Ablation Examples} \label{app:sec:addvis.ablations}

\textbf{Masked Depth and Disocclusion.}
We show an additional experimental ablation example in Fig.~\ref{fig:ablation.duck}, removing masked depth supervision and disocclusion handling (as in Fig.~\ref{fig:ablation}).
Removing the former causes significantly damaged geometry (and thus considerable visual artifacts as well), while ablating the latter increases blurriness in the disoccluded region (i.e., around newly unveiled details near the occlusion boundary).

\textbf{Disparity Smoothing.}
In Fig.~\ref{fig:smoothing.ablation}, 
    we consider the effect of ablating our disparity smoothing approach (see \autoref{sec:depth.alignment} and \autoref{app:sec:disp.smoothing}),
    utilized for obtaining depth in the masked area 
        and matching it to the surrounding scene geometry.
    Particularly close to the mask boundary, 
        we see that the \textit{un}smoothed geometry has a much more jarring transition
        between the masked and unmasked areas.

\subsection{Qualitative Results} \label{app:sec:addvis.qual}

\textbf{Comparisons.}
Additional comparisons to SPIn-NeRF, NeRF-In, and DreamFusion are shown for novel view synthesis in Fig.~\ref{fig:baseline.qualitative.nerfin.compare}.
Notice that utilizing the DreamFusion~\cite{poole2022dreamfusion} loss along with the Masked-NeRF (see~\autoref{sec:experiments} and \autoref{sec:app:baselines}) can result in unrealistic colours (first row) and sometimes a failure to converge (second row), though the quality improves over Masked-NeRF alone (see Table~\ref{tab:multiview.inpainting}).
NeRF-In~\cite{nerf.in} is blurry in masked areas, as the textures do not match well in a pixelwise manner.
SPIn-NeRF~\cite{spinnerf} reduces this blurriness considerably, but still incurs some level of blur, especially in the presence of more complex textures (e.g., second row).
In contrast, our method provides sharp details for all cases. 

\textbf{Controllability.}
We also provide more examples of controllable inpainting in Fig.~\ref{fig:controllable.second}.
Notice that we can easily control various aspects of the inpainted scene, such as the presence or absence of roots in the tree (upper rows) or the length of the stone bench (lower rows), by simply changing the inpainting of the single reference image. 
For additional examples of controllable insertion,
see also Fig.~\ref{fig:controllable}.

\section{Implementation Details} \label{sec:implementation.details}

In our experiments, both $N_\text{depth}$ and $N_\text{bilateral}$ are set to $2000$. We train each scene for $10000$ iterations. The disoccusion handling is run every $N_\text{do} = 3000$ iterations. The weights $\gamma_\text{depth}^\text{masked}$, $\gamma_\text{rec}^\text{masked}$, $\gamma_\text{do}$, $\eta_\text{do}$, and $\gamma_\text{smooth}$ are set to $4$, $2$, $1$, $0.25$, and $1000$, respectively, and $c_\text{ei}$ is set to $2$. We follow~\cite{spinnerf} and use a combination of \cite{instant.ngp} and \cite{ds.nerf} for faster convergence, and dilate all of the masks for 5 iterations with a $5\times5$ kernel to make sure that the masks cover the whole object, and to mask some of the shadows of the unwanted object. All of the images are downsized four times to reduce memory usage and match the experiments of SPIn-NeRF~\cite{spinnerf}. We also use the distortion loss proposed by~\cite{mipnerf.360} for reducing the floater artifacts. We set the weight of the distortion loss to $0.01$. For generating multiple inpainted source views, we leverage the diversity of denoising diffusion models, and use stable-diffusion inpainting v2~\cite{stable.diffusion}. For inpainting the residuals with the bilateral solver, we set the brightness and colour bandwidths to $4$, while the spatial bandwidth was set to $128$. The strength smoothness and the number of PCG iterations are set to $128$ and $25$, respectively. 
For disocclusion handling, we use LaMa \cite{lama} as the 2D inpainter and use three target images for $T$ (corresponding to the cameras furthest leftward, rightward, and upward). A small morphological dilation (four iterations with a $3\times 3$ kernel) is applied to remove noise from the disocclusion masks. The bilateral filter in the disocclusion case uses a spatial bandwidth of only 8. Our implementation is mainly in PyTorch~\cite{pytorch}. For generating the inpaintings for \textit{Ours-SD}, we used stable diffusion inpainting v2~\cite{stable.diffusion}, and a simple per-scene text prompt describing the inpainted scene. Below are the text prompts used for SPIn-NeRF scenes:
\begin{itemize}
    \item A stone bench, a bush in the background, the bench is grey with a rectangular shape in perspective, photorealistic 8k
    \item A wooden tree trunk on dirt, photorealistic 8k
    \item A red fence, photorealistic 8k
    \item Stone stairs, photorealistic 8k
    \item A circular lid made of rusty iron on a grass ground, photorealistic 8k
    \item A corner of a brick wall, photorealistic 8k
    \item A wooden bench in front of a white fence, photorealistic 8k
    \item An image of nature with grass, bushes in the background, photorealistic 8k
    \item A desk in front of a brick wall with an iron pipe, photorealistic 8k
    \item A brick wall, photorealistic 8k
\end{itemize}
Note that we did not engineer the prompts to improve the results. We typically selected the first generated inpainting. However, as seen in Fig.~\ref{fig:different.sd.inpaintings}, sometimes, the stable diffusion inpainter adds objects in the scene; in those cases, we regenerated the output to get an inpainting without any additional object for a fair comparison to the baselines. For quantitative experiments, we always select the $30$-th image among the $60$ training views in SPIn-NeRF's dataset~\cite{spinnerf} as the reference view.

\section{Metrics: Additional Details} \label{app:sec:metrics}

The ill-posed nature of inpainting means that evaluation is non-trivial: 
    ``ground-truth'' images are merely one of an infinite number of possible solutions, any plausible member of which should be considered valid.
We therefore focus on evaluating perceptual quality and realism, rather than reconstruction, via two types of metrics:
    full-reference (FR) and no-reference (NR).

In the FR case, we utilize the ground-truth (GT) images of the scene without the object for comparison with LPIPS \cite{perceptual} and Frechet Inception Distance (FID) \cite{fid}.
LPIPS, a perceptual distance, is far more robust to changes that maintain textural consistency than pixelwise distances.
For FID, we compare the distributions of encoded statistics between the inpainted and GT images, 
    which confers high robustness to mismatches in local details, focusing instead on agreement in high-level visual appearance.
Both of these metrics were used previously for 3D inpainting evaluation \cite{spinnerf}. For both LPIPS and FID, we only perform the comparison inside the bounding boxes of the objects. We expand the bounding boxes by $10\%$ to match SPIn-NeRF's~\cite{spinnerf} setup. 

However, FR metrics are not completely robust to the choice of reference image, preferring solutions more similar to the GT over others that are equally perceptually realistic.
This is exacerbated if an inpainting model inserts new semantic content into a scene, as recent diffusion-based approaches are apt to do (e.g., \cite{stable.diffusion,ramesh2022hierarchical}), 
    whether it is perceptually realistic or not.
Thus, we consider two NR metrics, where image quality is assessed in a stand-alone manner.
The first measure is simply the variance of the image Laplacian, a classical measure of sharpness (e.g., \cite{pertuz2013analysis}),
    which has been previously used to evaluate 2D generative image models \cite{tolstikhin2018wasserstein,hirte2021realistic}.
The second is MUSIQ~\cite{ke2021musiq,pyiqa}, 
    a transformer-based model for 
    NR image quality assessment, meant to reproduce human perceptual judgments. 

Note that our metrics in the FR case are computed against bounding boxes (containing the object mask) in \textit{held-out} views, 
    while our NR sharpness metrics are computed across 120 renders from a camera in a spiralling pattern (in video form).
In this way, we assess inpainting quality in its full 3D context; i.e., we ensure that the inpainting quality generalizes to novel views.

\end{document}